\documentclass[sigconf]{acmart}
\usepackage{multirow}
\usepackage{booktabs}
\usepackage{graphicx}
\usepackage{array}
\usepackage{algorithm}
\usepackage{algpseudocode}
\usepackage{tcolorbox}
\usepackage{color, colortbl}
\usepackage{amsmath}
\usepackage{xspace}
\usepackage{enumitem}
\usepackage{hyperref}
\usepackage{listings}
\usepackage{pifont}
\usepackage{soul}        
\usepackage{xcolor}      
\usepackage{url}
\usepackage{subfigure}

\definecolor{cadmiumgreen}{rgb}{0.0, 0.42, 0.24}
\definecolor{codehlcolor}{rgb}{0.6, 1.0, 0.6}
\definecolor{codehlerrcolor}{rgb}{1.0, 0.6, 0.6}
\definecolor{codehlseederrcolor}{rgb}{1.0, 0.85, 0.4}

\newcommand\approach{{MGDebugger}\xspace}
\newcommand{\codeurl}{\url{https://github.com/YerbaPage/MGDebugger}}
\copyrightyear{2026}
\acmYear{2026}
\setcopyright{cc}
\setcctype{by}
\acmConference[ICSE '26]{2026 IEEE/ACM 48th International Conference on Software Engineering}{April 12--18, 2026}{Rio de Janeiro, Brazil}
\acmBooktitle{2026 IEEE/ACM 48th International Conference on Software Engineering (ICSE '26), April 12--18, 2026, Rio de Janeiro, Brazil}
\acmPrice{}
\acmDOI{10.1145/3744916.3773255}
\acmISBN{979-8-4007-2025-3/26/04}

\newcommand{\edit}[1]{{\color{black}#1}}

\title{From Code to Correctness: Closing the Last Mile of Code Generation with Hierarchical Debugging}

\author{Yuling Shi}
\affiliation{%
  \institution{Shanghai Jiao Tong University}
  \city{Shanghai}
  \country{China}
}
\email{yuling.shi@sjtu.edu.cn}

\author{Songsong Wang}
\affiliation{%
  \institution{University of California, Davis}
  \city{Davis}
  \country{USA}
}
\email{ssswang@formerstudents.ucdavis.edu}

\author{Chengcheng Wan}
\affiliation{%
  \institution{East China Normal University}
  \city{Shanghai}
  \country{China}
}
\email{ccwan@sei.ecnu.edu.cn}

\author{Min Wang}
\affiliation{%
  \institution{University of Pennsylvania}
  \city{Philadelphia}
  \country{USA}
}
\email{minyun@seas.upenn.edu}

\author{Xiaodong Gu}
\authornote{Corresponding author.}
\affiliation{%
  \institution{Shanghai Jiao Tong University}
  \city{Shanghai}
  \country{China}
}
\email{xiaodong.gu@sjtu.edu.cn}

\begin{document}

\begin{abstract}
    While large language models have made significant strides in code generation, the pass rate of the generated code is bottlenecked on subtle errors, often requiring human intervention to pass tests, especially for complex problems. Existing LLM-based debugging systems treat generated programs as holistic units, failing to address bugs at multiple levels of granularity, from low-level syntax errors to high-level algorithmic flaws. In this paper, we introduce \textit{Multi-Granularity Debugger} (\approach), a hierarchical code debugger that isolates, identifies, and resolves bugs at various levels of granularity. \approach decomposes problematic code into a hierarchical tree structure of subfunctions, with each level representing a particular granularity of error. During debugging, it analyzes each subfunction and iteratively resolves bugs in a bottom-up manner. To effectively test each subfunction, we propose an LLM-simulated Python executor, which traces code execution and tracks important variable states to pinpoint errors accurately. 
    Extensive experiments with both open-source and commercial LLMs demonstrate that \approach significantly outperforms existing debugging systems, achieving up to 18.9\% improvement in accuracy over seed generations in HumanEval and a 97.6\% repair success rate in HumanEvalFix. Furthermore, \approach effectively generalizes to real-world software defects, fixing 129 bugs in Defects4J v1.2 and 64 bugs in Defects4J v2.0, outperforming state-of-the-art program repair approaches by 13.2\% and 8.5\% respectively. These results demonstrate \approach's robustness and effectiveness across different scenarios, establishing it as a powerful tool for closing the last mile of code generation.\footnote{Code and data available at \codeurl}
\end{abstract}

\maketitle
\section{Introduction}

Large language models (LLMs) such as GPT-4~\citep{openai2023gpt4}, LLaMA~\citep{touvron2023llama}, and DeepSeek-Coder~\citep{zhu2024deepseekcoderv2} have made significant advances in AI-assisted coding tasks~\citep{chen2021evaluating,li2022competitionlevel,just2014defects4j,shi2024between}.
Trained on vast corpora of text and code, LLMs can understand and generate code snippets for various programming tasks, ranging from simple data structures to complex algorithmic problems~\citep{li2022competitionlevel,wang2025dissect}. These models have demonstrated proficiency in tasks such as code completion, bug detection, and even tackling competitive programming challenges~\citep{chen2021evaluating,zhang2024codeagent,wen2024learning}. 

While the code generated by large language models generally meets the requirements, it often contains critical errors that require human intervention to pass tests~\citep{liu2023your,dou2024what}. This has gradually led to a new development paradigm: large models generate the code, while humans fix it. Therefore, the ``last mile'', as well as the most crucial step, of code generation is how to efficiently repair the code generated by large models.

Numerous efforts have been made to debug LLM-generated code. The most popular way is to reuse the LLM generator to debug the generated code with the feedback from test case execution~\citep{chen2023teaching,zhong2024debug,hu2024leveraging}.
While these methods increase the pass rates, they treat the erroneous program as a holistic set of statements \edit{(i.e., as a single unit without considering the internal structure or modularity)}~\citep{chen2023teaching,shinn2023reflexion,zhong2024debug,ding2024cycle} regardless of the varying types and levels of failures. Failures of test cases arise from different levels of factors, from low-level syntactic errors to high-level algorithmic flaws. A holistic treatment overlooks the internal structure of the code and limits the effectiveness of the debugging systems, especially when dealing with complex programs that need debugging across different modules~\citep{zeller2009why,tian2024debugbench}. 

In this paper, we introduce Multi-granularity Debugger (\approach), a novel debugging method for LLM-generated code.
Instead of treating entire functions as single units, \approach employs a hierarchical, bottom-up strategy to systematically debug code. It begins by decomposing the code into a tree structure of sub-functions, allowing for the isolation of semantic units for independent debugging. Each sub-function is debugged progressively, starting with the most granular ones and working upward to higher-level compositions until the entire code is repaired. To effectively test and debug each subfunction, \approach generates test cases derived from the public test cases of the main function. Then, it employs an LLM-based execution simulator to track changes in key variables, facilitating precise and flexible error identification based on the failed test cases. 
Through debugging at multiple levels of granularity from the bottom up in a recursive manner, \approach can uncover and rectify bugs that traditional holistic debugging methods might overlook.

\begin{figure*}[t]
    \centering
    \includegraphics[width=0.8\linewidth]{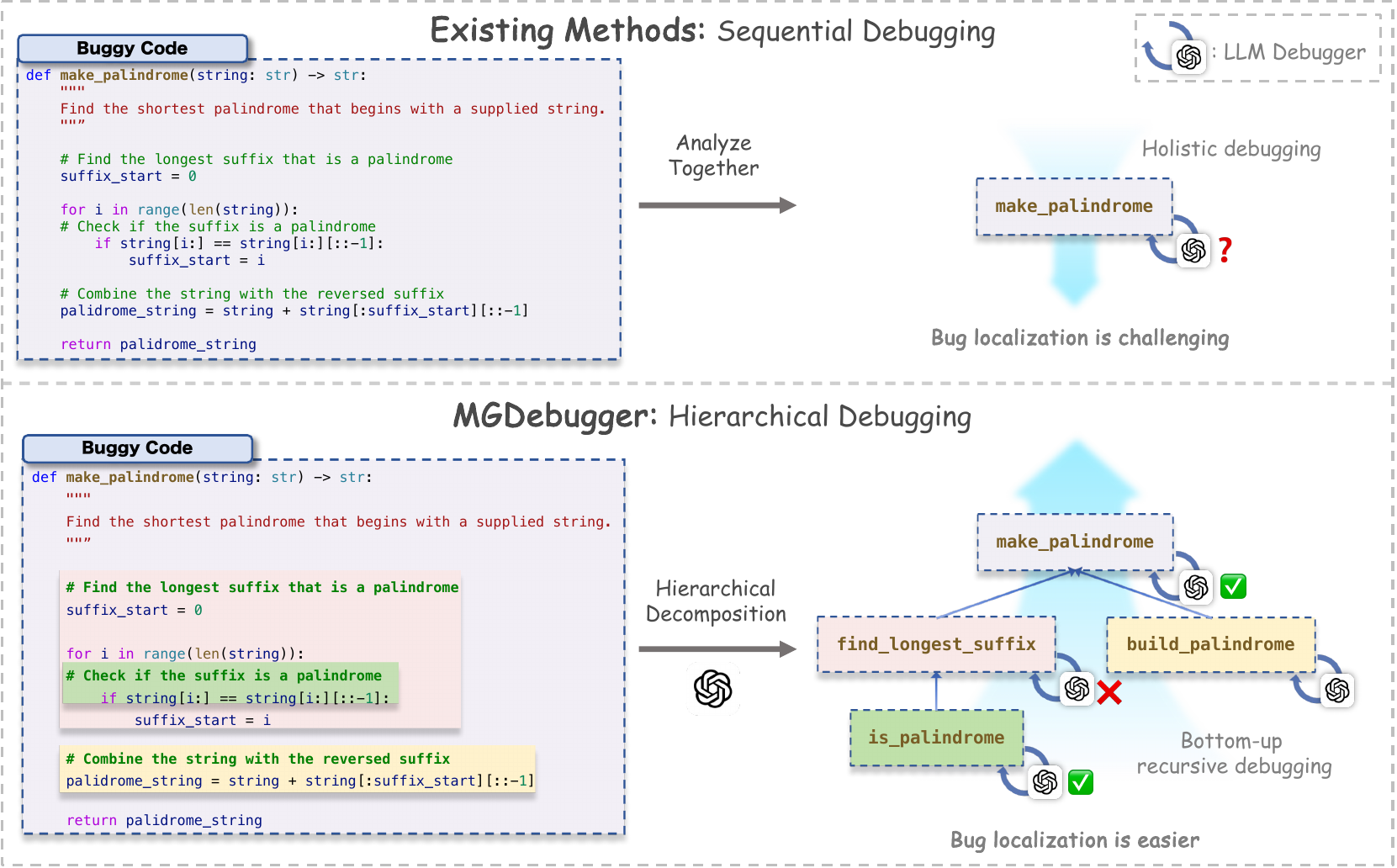}
    \vspace{-0.2cm}
    \caption{Workflow of \approach compared to existing methods.}
    \label{fig:overview}
    \vspace{-0.2cm}
\end{figure*}




Extensive experiments across multiple models and benchmarks demonstrate that \approach significantly outperforms existing debugging methods. Using open-source models like DeepSeek-Coder-V2-Lite, CodeQwen1.5, and Codestral, \approach elevates accuracy from 75.6\% to 94.5\% on HumanEval~\citep{chen2021evaluating} and achieves a remarkable 97.6\% repair success rate on HumanEvalFix~\citep{muennighoff2023octopack}. The approach also generalizes well to state-of-the-art proprietary models, improving GPT-4o and Claude 3.5 Sonnet's performance by 6.1\% on HumanEval, reaching accuracies of 96.3\% and 95.7\% respectively.

Our comprehensive ablation studies confirm the vital role of each component in \approach, with the hierarchical debugging strategy being the most critical element. Removing this component leads to a significant drop in repair success rate from 76.3\% to 52.6\% on HumanEval, highlighting its importance for systematic bug identification and resolution. Detailed analyses across different bug categories reveal \approach's exceptional performance in handling various error types, achieving 100\% success rates in fixing missing logic, excess logic, operator misuse, variable misuse, and function misuse bugs with DeepSeek-Coder. Furthermore, \approach maintains robust performance across code snippets of varying lengths and complexities, consistently outperforming baseline methods especially for longer, more complex code where other approaches typically struggle.

While \approach is primarily designed to fix LLM-generated code and close the last mile of code generation, we also evaluate its generalizability to real-world software defects. When applied to the Defects4J dataset~\citep{just2014defects4j}, \approach successfully repairs 129 bugs in Defects4J v1.2 and 64 bugs in Defects4J v2.0, outperforming specialized program repair tools. This demonstrates that the hierarchical debugging principles with test case generation and LLM-simulated execution of \approach extend effectively beyond LLM-generated code to human-written software with complex project structures.

The main contributions of this work can be summarized as:
\begin{itemize}[leftmargin=12pt]
    \item We introduce \approach, a novel hierarchical debugging method to repair LLM-generated code at multiple levels of granularity.
    \item We propose an LLM-simulated execution approach that enables precise bug identification by tracing code execution and tracking variable states within subfunctions, eliminating the need for external debuggers.
    \item We demonstrate \approach's effectiveness through comprehensive experiments across multiple models and benchmarks, showing significant improvements over existing methods with up to 18.9\% accuracy gains on HumanEval and 97.6\% repair success rate on HumanEvalFix.
    \item We establish \approach's generalizability by successfully applying it to real-world software defects in Defects4J, outperforming specialized program repair tools and demonstrating its potential as a unified debugging framework for both AI-generated and human-written code.
\end{itemize}

\section{Methodology}

\subsection{Overview}


We present \approach, a novel bottom-up hierarchical debugging method for repairing LLM-generated code. The overall workflow of \approach transforming holistic debugging into hierarchical subfunction debugging is illustrated in Figure~\ref{fig:overview}, while the detailed debugging process for each subfunction is depicted in Figure~\ref{fig:subfunction_debug}. 


As shown in Figure~\ref{fig:overview}, \approach begins with \textit{Hierarchical Code Decomposition (Section~\ref{sec:decomposition})}, which decomposes the input buggy code into a hierarchical structure of subfunctions. This enables systematic identification and resolution of bugs at various levels of granularity. For each subfunction, \approach \textit{Generates Test Cases for Subfunctions (Section~\ref{sec:testcase})}, deriving private test cases from public test cases of the main function, as illustrated in Figure~\ref{fig:subfunction_debug}. \approach then executes these test cases and \textit{Debugs Subfunction with LLM-Simulated Execution (Section~\ref{sec:debugging})}. The LLM simulates step-by-step code execution for failed test cases, monitoring critical variables and state changes to pinpoint the cause of errors. Once a subfunction has been fixed, \approach updates it in the hierarchical structure and propagates the changes to dependent functions through \textit{Bottom-up Debugging (Section~\ref{sec:recursive})}. 
This hierarchical debugging approach not only tackles different types of bugs at various levels of abstraction but also guarantees a cohesive and systematic debugging process throughout the entire code structure.


\subsection{Hierarchical Code Decomposition}\label{sec:decomposition}


Modularizing and decomposing complex code into smaller helper subfunctions has been proven to be helpful especially for large functions that are difficult to understand~\citep{jain2023llmassisted,zelikman2023parsel}. To enable hierarchical debugging, we need to transform the input code into a tree-like structure of subfunctions.

Specifically, given an LLM-generated function $f$, we decompose it into a hierarchical structure of subfunctions denoted as $(f_1,...,f_n)$. \edit{We formally define a "subfunction" as a contiguous sequence of statements $S=[s_1,s_2,...,s_k]$ within function $f$, such that $S$ represents a logically coherent and minimal unit of functionality that can be abstracted as a function $f_i(x)$ with well-defined inputs and outputs.} These subfunctions can be organized as a tree $f_{\text{root}}=\text{TREE}(f_{\text{root}}, \text{CHILD}(f_{\text{root}})) $, where $ f_{\text{root}}$ represents the main function and $ \text{CHILD}(f)$ denotes the set of subfunctions directly called by $ f $. 
We leverage an LLM for the decomposition, adhering to three principles: (1) each subfunction represents the minimal reusable unit of code with a specific purpose, (2) higher-level functions call lower-level functions to achieve complex functionality, and (3) the overall structure facilitates isolated testing and debugging. 
As illustrated in Figure~\ref{fig:overview}, the resulting tree-like structure allows us to isolate logical units of the code, enabling more focused debugging efforts across different levels of granularity~\citep{woodfield1981effect,isazadeh2017source}. 
\edit{The prompt template for hierarchical decomposition is shown below:

\begin{tcolorbox}[title=Prompt Template for Hierarchical Decomposition, fontupper=\footnotesize, fontlower=\footnotesize, top=1mm, bottom=1mm, left=1mm, right=1mm, toptitle=0.5mm, bottomtitle=0.5mm]



Convert the following Python code into a tree-style hierarchical structure with multiple levels of sub-functions.
Each significant step or logical block should be its own function, and functions can call other sub-functions.
Ensure that the main function calls these sub-functions in the correct order, creating a tree-like structure.

Original Code:

\{code\}

Instruction:

Please first analyze the codes step by step, and then provide the converted code in a Python code block. When providing the final converted code, make sure to include all the functions in a flattened format, where each function is defined separately.
\end{tcolorbox}}




\subsection{Generating Test Cases for Subfunctions}\label{sec:testcase}

\begin{figure*}[t]
    \centering
    \includegraphics[width=0.95\linewidth]{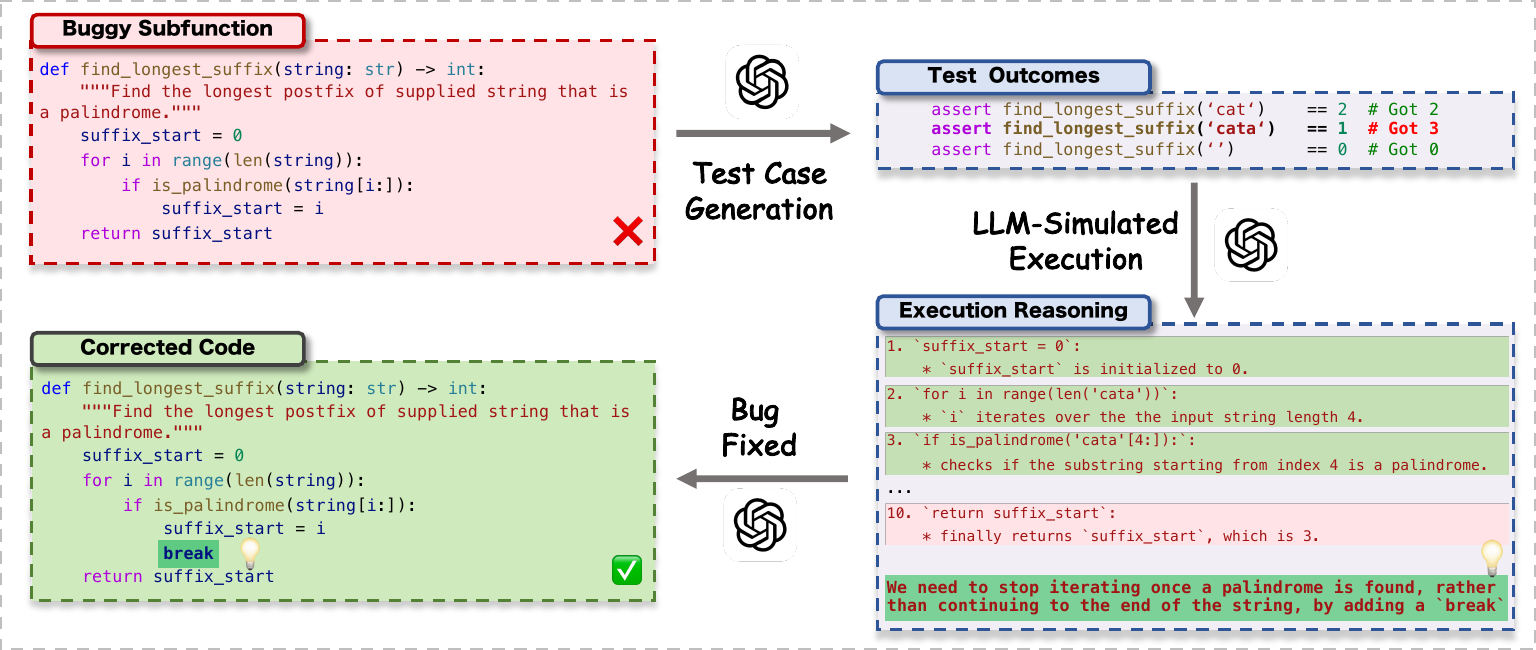}
    \vspace{-0.2cm}
    \caption{Illustration of the subfunction debugging process in \approach.}
    \label{fig:subfunction_debug}
    \vspace{-0.2cm}
\end{figure*}

Having obtained the hierarchy of subfunctions, we aim to verify the correctness of each subfunction. For this purpose, we generate test cases for each subfunction leveraging automatic unit test generation techniques~\citep{wang2021automatic,schafer2024empirical,liu2023rltf}. For each subfunction $ f_i\in f_\text{root}$, we generate a set of test cases $\mathcal{T}_i$. Following the problem settings from~\citet{chen2023teaching} and~\citet{zhong2024debug}, we assume that the public test cases for the main function $\mathcal{T}_{\text{pub}} $ have been provided, which is common in most code generation benchmarks~\citep{chen2021evaluating,hendrycks2021measuring,muennighoff2023octopack}\footnote{Otherwise, we can use LLM-generated test cases instead.}.
We can leverage these test cases to derive a set of corresponding test cases for each subfunction.

We employ the same LLM for the test case generation. For each $f_i\in f_\text{root}$. The LLM is now prompted to perform the following steps: (1) analyze how the subfunction is used within the main function and how it contributes to the expected outputs in the public test cases; (2) for each public test case, reason through the overall code structure step by step to figure out the input and expected output for the subfunction. This approach ensures that the generated test cases are not only reflective of the subfunction's intended functionality but also contextualized within the constraints provided by the public test cases, enhancing the robustness and relevance of the test cases. 
\edit{The prompt template for test case generation is shown below:

\begin{tcolorbox}[title=Prompt Template for Test Case Generation, fontupper=\footnotesize, fontlower=\footnotesize, top=1mm, bottom=1mm, left=1mm, right=1mm, toptitle=1mm, bottomtitle=1mm]
SYSTEM\_PROMPT:

You are an AI assistant specialized in analyzing Python functions and generating test cases.

USER\_PROMPT:

Full Code:

\{full\_code\}

Public Test Cases for the Main Function:

\{public\_test\_cases\}

Instruction:

Please analyze how the \{function\_name\} function is used within the main function and how it contributes to the expected outputs in the gold test cases. For each test case, you should analyze step-by-step based on both the input and the expected output of the main function, and then provide the corresponding input and expected output for the \{function\_name\} function. Ensure that the generated test cases are consistent with the behavior expected in the public test cases.

\end{tcolorbox}} 

\subsection{Debugging Subfunctions with LLM-simulated Execution}\label{sec:debugging}

With the generated test cases, we debug each subfunction by running them on the test case inputs, obtaining the results, and comparing these results against the expected outcomes in the test cases. When a failed test case is identified, we fix the corresponding subfunction and produce a corrected version.

One straightforward way to implement this process is to use an external Python executor to monitor runtime variable values~\citep{zhong2024debug}. However, when debugging high-level functions, tracking variable values within lower-level subfunctions is often unnecessary, as their correctness is ensured by the bottom-up debugging methodology. Furthermore, directly collecting all execution traces from the external debugger can add unnecessary overhead and complexity to the process.

Inspired by the methodology in~\citet{li2023chain}, we propose an LLM-simulated code executor, which prompts LLM to act as a code interpreter and track the code execution. \edit{We choose LLM-simulated execution over real execution for several key reasons~\cite{la2024code,ni2024next,zheng2024opencodeinterpreter}: (1) Many code snippets require specific environments, hardware, or dependencies that are difficult to set up or may not be available (e.g., embedded systems, special libraries), (2) Some code cannot be executed due to compilation errors or security concerns, necessitating sandboxing or virtual execution.}
As shown in Figure~\ref{fig:subfunction_debug}, we request the LLM to simulate the execution process, reasoning about key variables and their states at each step, and thoroughly analyzing the failed test cases. This eliminates the need for an external debugger, offering a more flexible and efficient debugging solution. In addition, the LLM can accurately identify where errors occur and grasp their surrounding context. 
\edit{The prompt template for debugging subfunction is shown below:

\begin{tcolorbox}[title=Prompt Template for Debugging Subfunction, fontupper=\footnotesize, fontlower=\footnotesize, top=1mm, bottom=1mm, left=1mm, right=1mm, toptitle=1mm, bottomtitle=1mm]
SYSTEM\_PROMPT:

You are an AI assistant helping to debug Python functions.

USER\_PROMPT:

Debug the following Python function. The function is not passing all test cases. Analyze the code, identify the bug, and provide a fixed version of the function.

Function Code:

\{function\_code\}

Test Case Results:

\{test\_case\_results\}

Instruction:

Please try to work as a Python interpreter to execute the code step-by-step. Identify the change of each variable as you ``run'' the code line-by-line. Based on the execution trace, try to identify the bug and provide the final fixed code in a Python code block.

\end{tcolorbox}}


\begin{algorithm}[ht]
    \caption{\approach: Bottom-up Recursive Debugging}
    \label{alg:approach}
    \begin{algorithmic}[1]
        \Statex \textbf{Input:} $f$: Input LLM-generated function; $\mathcal{T}_{\text{pub}}$: Public test cases.
        
        \Statex \textbf{Output:} $f'$: Debugged $f$.

        \Function{\approach}{$f, \mathcal{T}_{\text{pub}}$}
            \If{$f$ has subfunctions $\{f_1, \dots, f_n\}$}
                \For{$f_i\in f$} \Comment{Depth-first traversal}
                    \State $f_i' \gets \textsc{\approach}(f_i, \mathcal{T}_{\text{pub}})$ \Comment{Recursive debugging}
                    \State $f_i$ = $f_i'$ \Comment{Replace $f_i$ with the debugged version}
                \EndFor
            \EndIf
            
            \State $\mathcal{T}_f \gets \textsc{GenTest}(f, \mathcal{T}_{\text{pub}})$ \Comment{Generate test cases for $f$}
            \State $\mathcal{R}_f \gets \textsc{Exec}(f, \mathcal{T}_f)$ \Comment{Execute test cases for $f$}
            
            \If{pass ($\mathcal{R}_f$, $\mathcal{T}_f$)}
                \State \Return $f$ \Comment{Correct function; keep as is}
            \Else
                \State $f' \gets \textsc{Debug}(f, \mathcal{T}_f, \mathcal{R}_f)$ \Comment{Debug function $f$ based on test results $\mathcal{R}_f$}
                \State \Return $f'$ \Comment{Return the corrected code}
            \EndIf
        \EndFunction
    \end{algorithmic}
\end{algorithm}

\subsection{Bottom-up Debugging}\label{sec:recursive}

Having introduced code decomposition and the debugging process for each subfunction, we now outline the overall debugging workflow.

We initiate the process by calling \approach on the main function with the decomposed code $f_\text{root}$ and the set of public test cases $\mathcal{T}_{\text{pub}}$. \approach traverses the hierarchical structure in a depth-first manner, recursively debugging each subfunction before moving on to the higher-level functions. For each specific subfunction, \approach generates relevant test cases and debugs the function based on the results. When a fix is identified, \approach updates the function and propagates the changes to the dependent functions. This recursive, bottom-up strategy systematically addresses bugs by beginning with the most granular levels and progressively advancing through the function hierarchy. This method accommodates various types of bugs at different abstraction levels, from low-level syntax errors to high-level logical flaws, by focusing on one level of the hierarchy at a time and building up the corrected code in a structured manner. The detailed algorithm is presented in Algorithm~\ref{alg:approach}.

\section{Experiments}

We conduct experiments to evaluate the effectiveness of \approach, aiming to answer the following research questions:
\begin{itemize}[leftmargin=12pt]
    \item \textbf{RQ1:} How effective is \approach in debugging LLM-generated code?
    \item \textbf{RQ2:} To what extent do individual components of \approach contribute to its overall performance?
    \item \textbf{RQ3:} How does \approach perform in fixing bugs across different categories?
    \item \textbf{RQ4:} How consistent is \approach's performance across different debugging scenarios?
    \item \textbf{RQ5:} How will \approach generalize to fixing real-world software defects?
\end{itemize}

\subsection{Setup}

\textbf{Models}
We select three state-of-the-art open-source LLMs ranging from 7B to 22B parameters as backbones for code generation and debugging in the experiments: CodeQwen1.5 (7B)~\citep{bai2023qwen}, DeepSeek-Coder-V2-Lite (16B)~\citep{zhu2024deepseekcoderv2}, and Codestral (22B)~\citep{mistralaiteam2024codestral}. To validate the generalizability of \approach, stronger general-prupose models like GPT-4o~\citep{openai2023gpt4}, Claude 3.5 Sonnet~\cite{anthropic2024introducing}, and LLaMA 3.1 70B~\citep{touvron2023llama} are also included in the Experiments.

\textbf{Datasets}
We conduct experiments on three widely used benchmarks for LLM-based code generation and debugging.
HumanEval~\citep{chen2021evaluating} and MBPP~\citep{austin2021program} are two widely used benchmarks for evaluating code generation systems with 164 and 500 problems, respectively. 
The HumanEvalFix dataset~\citep{muennighoff2023octopack} consists of 164 buggy functions with six different bug categories: value misuse, missing logic, excess logic, operator misuse, variable misuse, and function misuse.

\textbf{Metrics}
We adopt two metrics to evaluate our method: 1) \textit{Accuracy}~\citep{chen2023teaching,zhong2024debug}, which measures the overall proportion of correct code samples among all generated code samples after debugging. A code is correct iff it passes all private test cases assigned to it. 2) \textit{Repair Success Rate} (RSR)~\citep{yasunaga2021breakitfixit}, which refers to the proportion of fixed code samples to the total number of buggy code samples. The former metric indicates to which extent \approach can improve the code generation systems, and the latter metric directly reflects the effectiveness of \approach in fixing bugs from LLM-generated code. 

\textbf{Baselines}
We compare \approach with eight state-of-the-art methods for debugging LLM-generated code.
\edit{0)~\textit{No-Debugging} baseline refers to directly using the LLM to generate code without applying any debugging method, following~\citep{chen2023teaching,zhong2024debug}. This baseline creates the initial set of seed programs, the subset of buggy programs then become targets for various debugging methods.}
1)~\textit{Simple Feedback} is a basic baseline that informs the LLM that the code is incorrect and asks it to fix the issue.
2)~\textit{Self-Edit}~\citep{zhang2023selfedit} prompts the LLM to edit the code based on the execution results of the test cases.
3)~\textit{Self-Debugging}~\citep{chen2023teaching} has two variants: \textit{Self-Debugging (Expl.)} prompts the LLM to explain the generated code line-by-line, while \textit{Self-Debugging (Trace)} asks the LLM to dry run the code for debugging.
4)~\textit{LDB}~\citep{zhong2024debug} segments the code into basic \textit{blocks}, \textit{functions} or \textit{lines}, and tracks variable values during runtime after each block to verify correctness against the task description.
5)~\textit{Reflexion}~\citep{shinn2023reflexion} asks the LLM to reflect on the previous code given execution results and uses a memory buffer to enable iterative refinement.

\textbf{Implementation Details}
We generate seed programs for HumanEval and MBPP using the BigCode Evaluation Harness framework\footnote{\url{https://github.com/bigcode-project/bigcode-evaluation-harness}}. The specific versions of models used in our experiments are DeepSeek-Coder-V2-Lite-Instruct, CodeQwen1.5-7B-Chat, and Codestral-22B-v0.1. All experiments are conducted on NVIDIA A100 GPUs with 80GB memory. During debugging, we use the vLLM engine\footnote{\url{https://github.com/vllm-project/vllm}} to serve the LLMs, setting the maximum token length according to each LLM's max length. Following~\citet{zhong2024debug}, we limit the maximum number of debugging iterations to 10 for all methods on HumanEval, MBPP, and HumanEvalFix. To obtain visible test cases for HumanEval and HumanEvalFix, we extract the given visible test cases from the task description. For MBPP, we use the first test case of each problem as the visible test case and use the rest as hidden test cases, in line with the settings referenced from~\citet{chen2023teaching} and~\citet{zhong2024debug}. Additionally, \edit{the sampling temperature that controls the randomness of LLM decoding} is set to 0.8 in \approach. \edit{For baseline methods that incorporate external feedback from the environments, we follow~\cite{zhong2024debug} to execute the complete test code to collect runtime feedback.}


\subsection{Main Results (RQ1)}

\begin{table}[t]
    \centering
    \renewcommand{\arraystretch}{1.05}
    \caption{Main results.}
    \resizebox{\linewidth}{!}{
    \begin{tabular}{l@{\hspace{1.5mm}}c@{\hspace{1.5mm}}c@{\hspace{1.5mm}}c@{\hspace{1.5mm}}c@{\hspace{1.5mm}}c@{\hspace{1.5mm}}c}
    \toprule
    \multirow{3}[3]{*}{\textbf{Method}} & \multicolumn{6}{c}{\textbf{Dataset}} \\
    \cmidrule{2-7} 
    & \multicolumn{3}{c}{\textbf{HumanEval}} & \multicolumn{3}{c}{\textbf{MBPP}} \\
    \cmidrule(lr){2-4} \cmidrule(lr){5-7}
    & \textbf{Acc. (\%)} & \textbf{RSR (\%)} & & \textbf{Acc. (\%)} & \textbf{RSR (\%)} & \\
    \midrule
    \rowcolor[HTML]{EFEFEF} 
    \multicolumn{7}{c}{DeepSeek-Coder-V2-Lite} \\
    \hline
    No-Debugging & 76.8 & -- & & 67.2 & -- & \\
    Simple Feedback & 82.3 (\textcolor{cadmiumgreen}{+5.5}) & 23.7 & & 69.4 (\textcolor{cadmiumgreen}{+2.2}) & 6.7 & \\
    Self-Edit & 82.9 (\textcolor{cadmiumgreen}{+6.1}) & 26.3 & & 71.2 (\textcolor{cadmiumgreen}{+4.0}) & 12.2 & \\
    LDB (Block) & 84.1 (\textcolor{cadmiumgreen}{+7.3}) & 31.6 & & 74.0 (\textcolor{cadmiumgreen}{+6.8}) & 20.7 & \\
    LDB (Line) & 82.3 (\textcolor{cadmiumgreen}{+5.5}) & 23.7 & & 71.8 (\textcolor{cadmiumgreen}{+4.6}) & 14.0 & \\
    LDB (Function) & 81.7 (\textcolor{cadmiumgreen}{+4.9}) & 21.1 & & 72.6 (\textcolor{cadmiumgreen}{+5.3}) & 16.5 & \\
    Self-Debugging (Expl.) & 87.2 (\textcolor{cadmiumgreen}{+10.4}) & 44.7 & & 73.4 (\textcolor{cadmiumgreen}{+6.2}) & 18.9 & \\
    Self-Debugging (Trace) & 86.0 (\textcolor{cadmiumgreen}{+9.2}) & 39.5 & & 72.6 (\textcolor{cadmiumgreen}{+5.3}) & 16.5 & \\
    Reflexion & 90.9 (\textcolor{cadmiumgreen}{+14.1}) & 60.5 & & 76.6 (\textcolor{cadmiumgreen}{+9.4}) & 28.7 & \\
    \textbf{\approach} & \textbf{94.5} (\textcolor{cadmiumgreen}{\textbf{+17.7}}) & \textbf{76.3} & & \textbf{80.0} (\textcolor{cadmiumgreen}{\textbf{+12.8}}) & \textbf{39.0} & \\
    \midrule
    \rowcolor[HTML]{EFEFEF} 
    \multicolumn{7}{c}{CodeQwen1.5} \\
    \hline
    No-Debugging & 76.2 & -- & & 67.4 & -- & \\
    Simple Feedback & 85.4 (\textcolor{cadmiumgreen}{+9.2}) & 38.5 & & 74.0 (\textcolor{cadmiumgreen}{+6.6}) & 20.2 & \\
    Self-Edit & 84.1 (\textcolor{cadmiumgreen}{+7.9}) & 33.3 & & 75.0 (\textcolor{cadmiumgreen}{+7.6}) & 23.3 & \\
    LDB (Block) & 79.3 (\textcolor{cadmiumgreen}{+3.1}) & 12.8 & & 72.8 (\textcolor{cadmiumgreen}{+5.4}) & 16.6 & \\
    LDB (Line) & 79.9 (\textcolor{cadmiumgreen}{+3.7}) & 15.4 & & 72.6 (\textcolor{cadmiumgreen}{+5.2}) & 16.0 & \\
    LDB (Function) & 80.5 (\textcolor{cadmiumgreen}{+4.3}) & 17.9 & & 72.8 (\textcolor{cadmiumgreen}{+5.4}) & 16.6 & \\
    Self-Debugging (Expl.) & 87.8 (\textcolor{cadmiumgreen}{+11.6}) & 48.7 & & 77.4 (\textcolor{cadmiumgreen}{+10.0}) & 30.7 & \\
    Self-Debugging (Trace) & 84.8 (\textcolor{cadmiumgreen}{+8.6}) & 35.9 & & 76.8 (\textcolor{cadmiumgreen}{+9.4}) & 28.8 & \\
    Reflexion & 87.8 (\textcolor{cadmiumgreen}{+11.6}) & 48.7 & & 78.6 (\textcolor{cadmiumgreen}{+11.2}) & 34.4 & \\
    \textbf{\approach} & \textbf{91.5} (\textcolor{cadmiumgreen}{\textbf{+15.3}}) & \textbf{64.1} & & \textbf{80.8} (\textcolor{cadmiumgreen}{\textbf{+13.4}}) & \textbf{41.1} & \\
    \midrule
    \rowcolor[HTML]{EFEFEF}
    \multicolumn{7}{c}{Codestral} \\
    \hline
    No-Debugging & 75.6 & -- & & 65.4 & -- & \\
    Simple Feedback & 88.4 (\textcolor{cadmiumgreen}{+12.8}) & 52.5 & & 71.6 (\textcolor{cadmiumgreen}{+6.2}) & 17.9 & \\
    Self-Edit & 86.0 (\textcolor{cadmiumgreen}{+10.4}) & 42.5 & & 75.8 (\textcolor{cadmiumgreen}{+10.4}) & 30.0 & \\
    LDB (Block) & 83.5 (\textcolor{cadmiumgreen}{+7.9}) & 32.5 & & 72.2 (\textcolor{cadmiumgreen}{+6.8}) & 19.7 & \\
    LDB (Line) & 83.5 (\textcolor{cadmiumgreen}{+7.9}) & 32.5 & & 71.8 (\textcolor{cadmiumgreen}{+6.4}) & 18.5 & \\
    LDB (Function) & 82.3 (\textcolor{cadmiumgreen}{+6.7}) & 27.5 & & 72.0 (\textcolor{cadmiumgreen}{+6.6}) & 19.1 & \\
    Self-Debugging (Expl.) & 89.6 (\textcolor{cadmiumgreen}{+14.0}) & 57.5 & & 76.4 (\textcolor{cadmiumgreen}{+11.0}) & 31.8 & \\
    Self-Debugging (trace) & 84.1 (\textcolor{cadmiumgreen}{+8.5}) & 35.0 & & 73.6 (\textcolor{cadmiumgreen}{+8.2}) & 23.7 & \\
    Reflexion & 86.6 (\textcolor{cadmiumgreen}{+11.0}) & 45.0 & & 75.2 (\textcolor{cadmiumgreen}{+9.8}) & 28.3 & \\
    \textbf{\approach} & \textbf{94.5} (\textcolor{cadmiumgreen}{\textbf{+18.9}}) & \textbf{77.5} & & \textbf{76.8} (\textcolor{cadmiumgreen}{\textbf{+11.4}}) & \textbf{32.9} & \\
    \bottomrule
    \end{tabular}%
    }
    \label{tab:main}%
\end{table}%


The results in Table~\ref{tab:main} show that \approach consistently outperforms existing approaches across all models and datasets. Specifically, \approach achieves the highest accuracy improvements, with gains of +15.3\% to +18.9\% on HumanEval and +11.4\% to +13.4\% on MBPP. These improvements are particularly notable when compared to baseline methods such as Self-Debugging (Expl.) and Reflexion, which also incorporate external feedback but exhibit lower gains in accuracy and RSR. The strong results across models of varying sizes highlight the adaptability of \approach to different LLM architectures.

Moreover, \approach demonstrates remarkable debugging capabilities, particularly with DeepSeek-Coder-V2-Lite (16B) and Codestral (22B), where it achieves an accuracy of 94.5\% on the HumanEval dataset, the highest score among all methods. This is especially impressive considering that \approach operates in a zero-shot setting without task-specific retraining. This result illustrates the inherent debugging ability of larger LLMs with \approach. Additionally, the method's performance on MBPP, achieving an RSR of up to 41.1\% with smaller models like CodeQwen1.5 (7B), further underscores its robustness. Statistical analysis using the Wilcoxon signed-rank test with 10 repetitions confirms that these improvements are significant (p < 0.005). In general, these results validate \approach as a highly effective and scalable debugging method for LLM-generated code. 

\begin{table}[t]
    \centering
    \caption{Results (Acc.) on HumanEval with advanced models.}
    \resizebox{\linewidth}{!}{
    \setlength{\tabcolsep}{2.5mm}{
    \begin{tabular}{lcccc}
    \toprule
    \multirow{2}{*}{\textbf{Method}} & \multicolumn{3}{c}{\textbf{Model}} \\
    \cmidrule(lr){2-4}
    & \textbf{GPT-4o} & \textbf{Claude 3.5 Sonnet} & \textbf{LLaMA 3.1 70B} \\
    \midrule
    No-Debugging & 90.2 & 89.6 & 79.3 \\
    Self-Debugging (Expl.) & 93.3 (\textcolor{cadmiumgreen}{+3.1}) & 93.3 (\textcolor{cadmiumgreen}{+3.7}) & 86.6 (\textcolor{cadmiumgreen}{+7.3}) \\
    Reflexion & 94.5 (\textcolor{cadmiumgreen}{+4.3}) & 93.9 (\textcolor{cadmiumgreen}{+4.3}) & 89.6 (\textcolor{cadmiumgreen}{+10.4}) \\
    \textbf{\approach} & \textbf{96.3} (\textcolor{cadmiumgreen}{\textbf{+6.1}}) & \textbf{95.7} (\textcolor{cadmiumgreen}{\textbf{+6.1}}) & \textbf{92.7} (\textcolor{cadmiumgreen}{\textbf{+13.4}}) \\
    \bottomrule
    \end{tabular}
    }
    }
    \label{tab:advanced_models}%
\end{table}

To further validate the effectiveness and generalizability of \approach across a wider spectrum of language models, we extended our evaluation to include recent advanced LLMs with strong natural language capabilities, including GPT-4o, Claude 3.5 Sonnet, and LLaMA 3.1 70B. We compare \approach with the most competitive methods, Self-Debugging (Expl.) and Reflexion, on the HumanEval dataset.

The results in Table~\ref{tab:advanced_models} demonstrate that \approach continues to outperform existing debugging approaches even on state-of-the-art proprietary models like GPT-4o and Claude 3.5 Sonnet, which already possess strong zero-shot code generation capabilities. With these advanced models, \approach achieves impressive accuracy improvements of +6.1\% for both GPT-4o and Claude 3.5 Sonnet, pushing their overall accuracy to 96.3\% and 95.7\% respectively on HumanEval. This represents a substantial improvement over strong baselines like Self-Debugging and Reflexion. Notably, the results on LLaMA 3.1 70B demonstrate that \approach provides even greater benefits for open-source models with enhanced natural language capabilities. The method yields a remarkable +13.4\% improvement in accuracy, bringing LLaMA 3.1 70B's performance to 92.7\%.

These findings strengthen our previous conclusions about the adaptability of \approach across different model architectures. We also reveal an interesting trend: while state-of-the-art models like GPT-4o and Claude 3.5 Sonnet start with higher baseline performance, \approach still provides substantial improvements, suggesting that even the most advanced LLMs benefit significantly from structured debugging approaches. This highlights the potential for \approach as a general enhancement method that can complement the evolving landscape of AI coding assistants.

\subsection{Ablation Study (RQ2)}
To understand the contribution of each component in \approach and validate our design choices, we conduct an ablation study by systematically removing key components of our method: hierarchical code decomposition, LLM-simulated execution, and test case generation for subfunction debugging. Additionally, we explore the impact of replacing simulated execution with real execution traces collected with LDB~\citep{zhong2024debug} to further validate our design choice. Each variant is evaluated on both the HumanEval and MBPP datasets using the DeepSeek-Coder-V2-Lite model.

\begin{table}[t]
\centering
\caption{Results of ablation study.}
\resizebox{\linewidth}{!}{
\setlength{\tabcolsep}{2.5mm}{
\begin{tabular}{lc@{}cc@{}ccc}
\toprule
\multirow{2}{*}{\textbf{Method}} & \multicolumn{2}{c}{\textbf{HumanEval}} & \multicolumn{2}{c}{\textbf{MBPP}} \\
\cmidrule(lr){2-3} \cmidrule(lr){4-5}
& \textbf{Acc. (\%)} & \textbf{RSR (\%)} & \textbf{Acc. (\%)} & \textbf{RSR (\%)} \\
\midrule
\textbf{\approach} & \textbf{94.5} (\textcolor{cadmiumgreen}{+17.7}) & \textbf{76.3} & \textbf{80.0} (\textcolor{cadmiumgreen}{+12.8}) & \textbf{39.0} \\
 -\space w/o Hierarchical Debugging & 89.0 (\textcolor{cadmiumgreen}{+12.2}) & 52.6 & 78.2 (\textcolor{cadmiumgreen}{+11.0}) & 33.5 \\
 -\space w/o Simulated Execution & 90.2 (\textcolor{cadmiumgreen}{+13.4}) & 61.3 & 79.2 (\textcolor{cadmiumgreen}{+12.0}) & 36.6 \\
 -\space w/o Test Case Generation & 90.9 (\textcolor{cadmiumgreen}{+14.1}) & 60.5 & 79.2 (\textcolor{cadmiumgreen}{+12.0}) & 36.6 \\
 -\space w/ Real Execution & 92.7 (\textcolor{cadmiumgreen}{+15.9}) & 47.4 & 79.0 (\textcolor{cadmiumgreen}{+11.8}) & 36.0 \\
 No-Debugging & 76.8 & -- & 67.2 & -- \\
\bottomrule
\end{tabular}
 }
 }
\label{tab:ablation}
\end{table}

\begin{table}[t]
    \centering
    \caption{Performance (RSR) on different bug categories in HumanEvalFix. The best and second-best scores are highlighted in bold and underline, respectively.}
    \resizebox{\linewidth}{!}{
    \begin{tabular}{l@{\hspace{1.5mm}}c@{\hspace{1.5mm}}c@{\hspace{1.5mm}}c@{\hspace{1.5mm}}c@{\hspace{1.5mm}}c@{\hspace{1.5mm}}c@{\hspace{1.5mm}}c}
    \toprule
    \textbf{Method} & \textbf{Value}\,\,\, & \textbf{Missing Logic}\,\,\, & \textbf{Excess Logic}\,\,\, & \textbf{Operator}\,\,\, & \textbf{Variable}\,\,\, & \textbf{Function} & \textbf{Overall} \\
    \midrule
    \rowcolor[HTML]{EFEFEF} 
    \multicolumn{8}{c}{DeepSeek-Coder-V2-Lite} \\
    \hline
    Simple Feedback & 84.9 & \underline{96.0} & 80.7 & 78.3 & \underline{86.4} & \underline{87.5} & 85.4 \\ 
    Self-Edit  & 78.8 & 92.0 & 80.7 & 82.6 & 84.1 & 62.5 & 82.3 \\ 
    LDB (Block) & 69.7 & \underline{96.0} & 74.2 & 87.0 & \underline{86.4} & 62.5 & 81.1 \\ 
    LDB (Line)  & 63.6 & 84.0 & 67.7 & 73.9 & 84.1 & 62.5 & 74.4 \\ 
    LDB (Function)  & 69.7 & 88.0 & 71.0 & 87.0 & 77.3 & 62.5 & 76.8 \\ 
    Self-Debugging (Expl.) & 66.7 & 80.0 & 64.5 & 78.3 & \underline{86.4} & 50.0 & 74.4 \\ 
    Self-Debugging (Trace)  & 81.8 & 88.0 & 71.0 & 78.3 & 79.6 & 75.0 & 79.3 \\ 
    Reflexion & \textbf{90.9} & \textbf{100.0} & \underline{90.3} & \underline{91.3} & \underline{86.4} & \textbf{100.0} & \underline{91.5} \\ 
    \textbf{\approach} & \underline{87.9} & \textbf{100.0} & \textbf{100.0} & \textbf{100.0} & \textbf{100.0} & \textbf{100.0} & \textbf{97.6} \\ 
    \midrule
    \rowcolor[HTML]{EFEFEF}
    \multicolumn{8}{c}{CodeQwen1.5} \\ 
     \hline
     Simple Feedback & \textbf{81.8} & \underline{92.0} & \underline{87.1} & 69.6 & 81.8 & \underline{87.5} & \underline{82.9} \\ 
     Self-Edit  & 72.7 & \underline{92.0} & 80.7 & 65.2 & \textbf{86.4} & \underline{87.5} & 80.5 \\ 
     LDB (Block) & 36.4 & 72.0 & 51.6 & 60.9 & 63.6 & 62.5 & 56.7 \\ 
     LDB (Line) & 36.4 & 76.0 & 45.2 & 56.5 & 54.6 & 50.0 & 52.4 \\ 
     LDB (Function) & 27.3 & 60.0 & 51.6 & 56.5 & 59.1 & 62.5 & 51.2 \\ 
     Self-Debugging (Expl.) & 69.7 & \underline{92.0} & \textbf{90.3} & 69.6 & 77.3 & 62.5 & 78.7 \\ 
     Self-Debugging (Trace) & 72.7 & 72.0 & 80.6 & 69.6 & 70.5 & 75.0 & 73.2 \\ 
     Reflexion & 66.7 & 88.0 & 80.6 & \underline{91.3} & \textbf{86.4} & 75.0 & 81.7 \\ 
     \textbf{\approach} & \underline{78.8} & \textbf{96.0} & \underline{87.1} & \textbf{95.7} & \underline{84.1} & \textbf{100.0} & \textbf{87.8} \\ 
    \midrule
    \rowcolor[HTML]{EFEFEF} 
    \multicolumn{8}{c}{Codestral} \\
    \hline
     Simple Feedback  & 75.8 & 92.0 & 67.7 & 82.6 & 84.1 & 62.5 & 79.3 \\ 
     Self-Edit & \underline{78.8} & \textbf{100.0} & 80.7 & \underline{87.0} & 84.1 & \textbf{87.5} & 85.4 \\ 
     LDB (Block)  & 66.7 & 92.0 & 67.7 & 82.6 & 81.8 & \textbf{87.5} & 78.1 \\ 
     LDB (Line)  & 63.6 & 92.0 & 64.5 & 82.6 & 81.8 & \underline{75.0} & 76.2 \\ 
     LDB (Function)  & 57.6 & 88.0 & 67.7 & \textbf{91.3} & 75.0 & \underline{75.0} & 74.4 \\ 
     Self-Debugging (Expl.)  & 75.8 & \underline{96.0} & \underline{83.9} & \underline{87.0} & \underline{90.9} & \textbf{87.5} & \underline{86.6} \\ 
     Self-Debugging (Trace)  & 57.6 & 84.0 & 64.5 & 73.9 & 81.8 & \underline{75.0} & 72.6 \\ 
     Reflexion & 69.7 & 88.0 & 61.3 & 82.6 & 88.6 & \underline{75.0} & 78.0 \\ 
     \textbf{\approach} & \textbf{87.9} & \textbf{100.0} & \textbf{87.1} & 82.6 & \textbf{95.5} & \underline{75.0} & \textbf{90.2} \\ 
    \bottomrule
    \end{tabular}%
    }
    \label{tab:humanevalfix}
\end{table}

As shown in Table~\ref{tab:ablation}, each component of \approach plays a crucial role in the overall effectiveness of the method. Among them, the hierarchical debugging strategy is the most impactful component. By ablating this strategy, the repair success rate drops significantly from 76.3\% to 52.6\% on HumanEval and from 39.0\% to 33.5\% on MBPP. This result highlights the importance of the hierarchical approach in systematically identifying and fixing bugs at different granularity levels. Additionally, the LLM-simulated execution and test case generation for subfunctions also facilitate debugging the decomposed code, yielding substantial improvements in accuracy and repair success rates. These results underscore the effectiveness of \approach's design choices and the importance of its hierarchical debugging strategy.

Furthermore, we observe that replacing simulated execution with real execution leads to a noticeable drop in performance, particularly in repair success rate (from 76.3\% to 47.4\% on HumanEval and from 39.0\% to 36.0\% on MBPP). This suggests that real execution traces, which include all detailed variable changes and execution steps, may introduce noise that distracts the model during debugging. In contrast, simulated execution allows \approach to focus on key variables and express debugging steps in a more reasoning-oriented manner, leading to better code understanding and more accurate repairs. Furthermore, simulated execution offers greater flexibility for extension to complex programming environments where execution environments may not be easily accessible~\citep{ni2024next}. This makes \approach more adaptable to a wider range of scenarios.

\subsection{Debugging Different Types of Bugs (RQ3)}



To assess the versatility and effectiveness of \approach across various bug categories, we carry out experiments using the HumanEvalFix dataset, which is specifically designed to evaluate code debugging performance. The dataset involves six distinct bug categories: value misuse, missing logic, excess logic, operator misuse, variable misuse, and function misuse, allowing us to examine how effectively \approach addresses different types of programming errors compared to existing methods. 

\begin{figure*}[t]
    \centering
    \subfigure[HumanEval dataset]{\label{fig:humaneval_code_length_effectiveness_tokens}
    \includegraphics[width=0.32\textwidth]{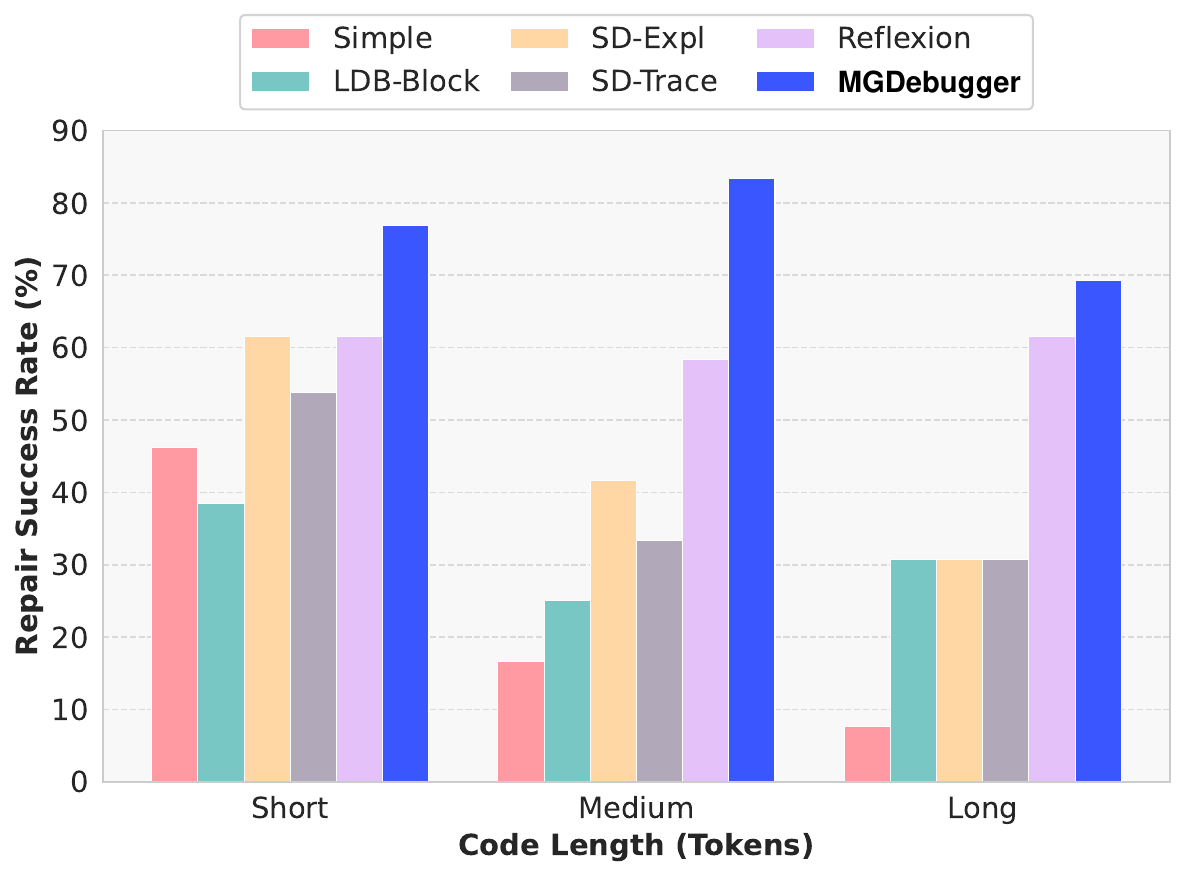}}
    \subfigure[MBPP dataset]{\label{fig:mbpp_code_length_effectiveness_tokens}
    \includegraphics[width=0.32\textwidth]{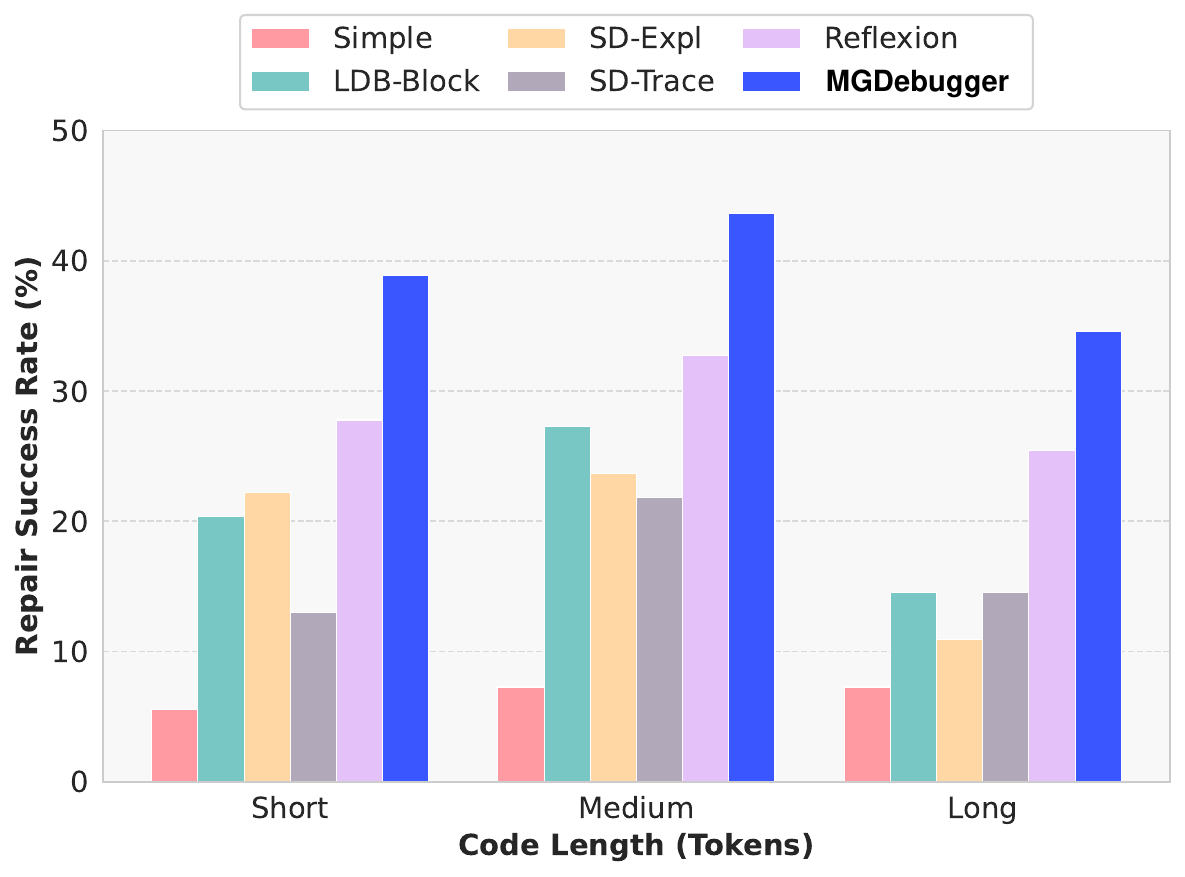}}
    \subfigure[HumanEvalFix dataset]{\label{fig:humanevalfix_code_length_effectiveness_tokens}
    \includegraphics[width=0.32\textwidth]{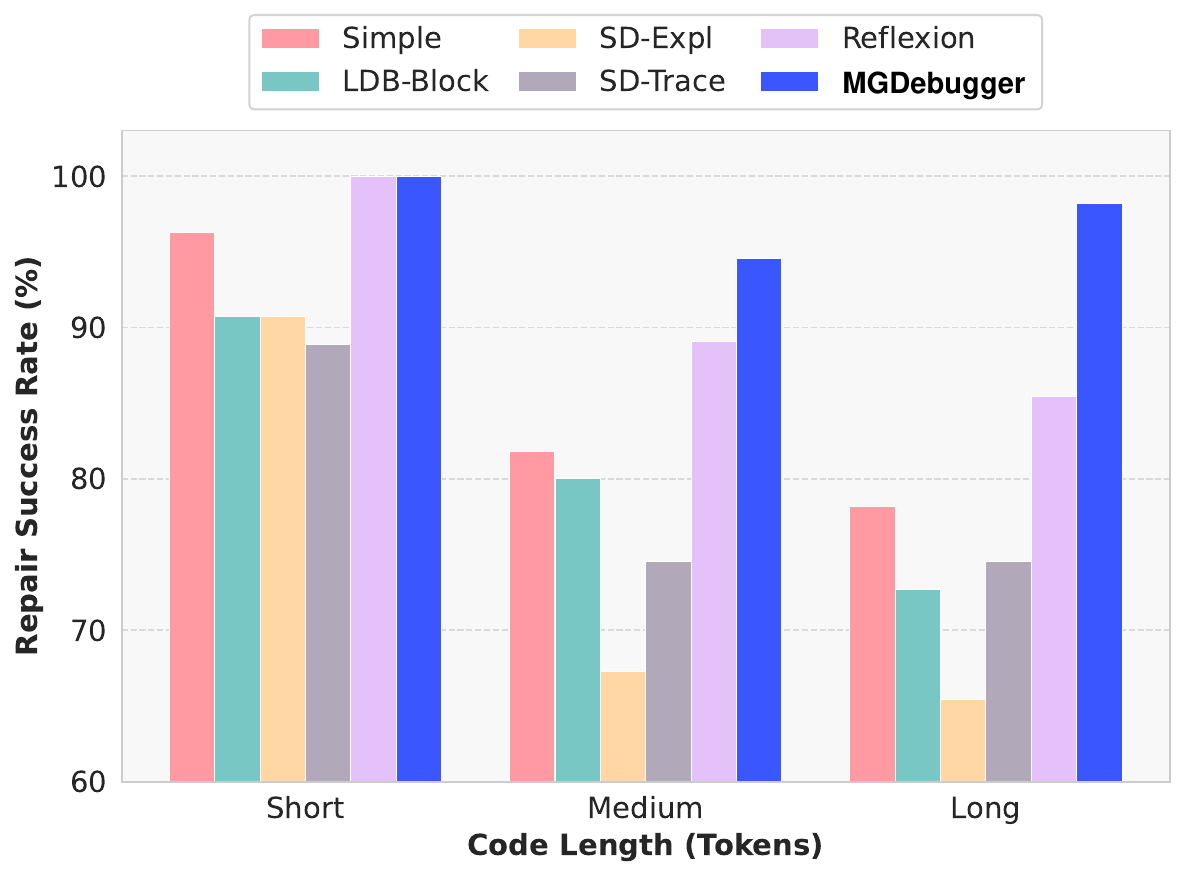}}
    \caption{Repair success rate of different methods when debugging code of different lengths.}
    \label{fig:code_length}
\end{figure*}

\begin{figure*}[t]
    \centering
    \subfigure[HumanEval dataset]{\label{fig:humaneval_retry_success_rate}
    \includegraphics[width=0.32\textwidth]{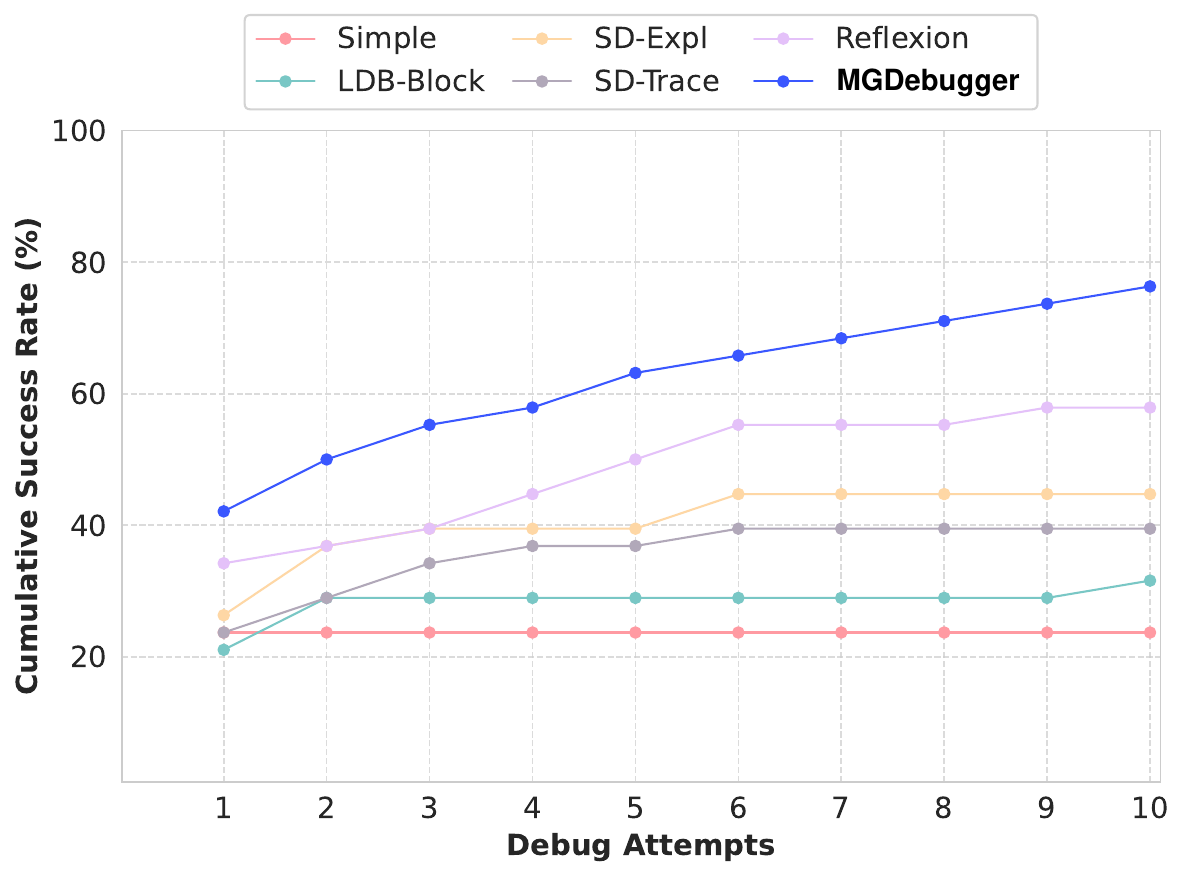}}
    \subfigure[MBPP dataset]{\label{fig:mbpp_retry_success_rate}
    \includegraphics[width=0.32\textwidth]{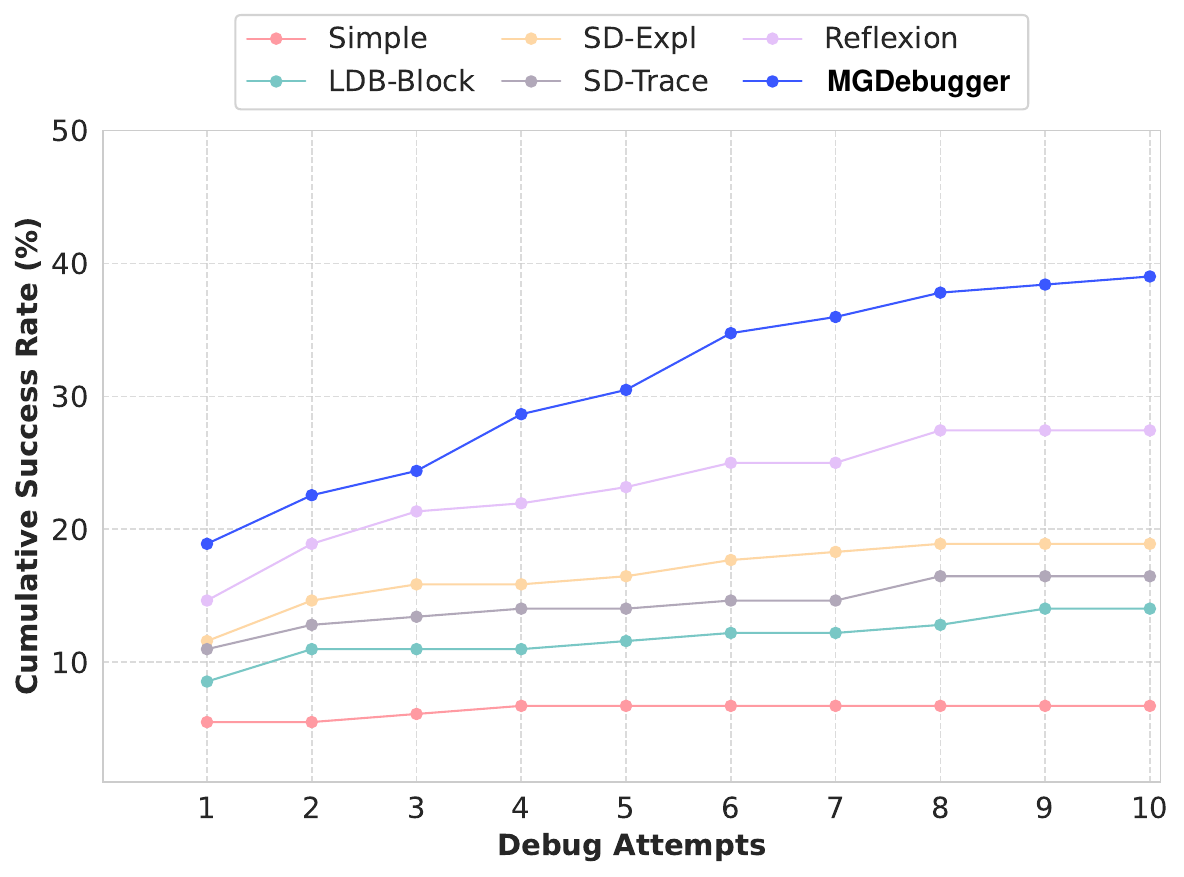}}
    \subfigure[HumanEvalFix dataset]{\label{fig:humanevalfix_retry_success_rate}
    \includegraphics[width=0.32\textwidth]{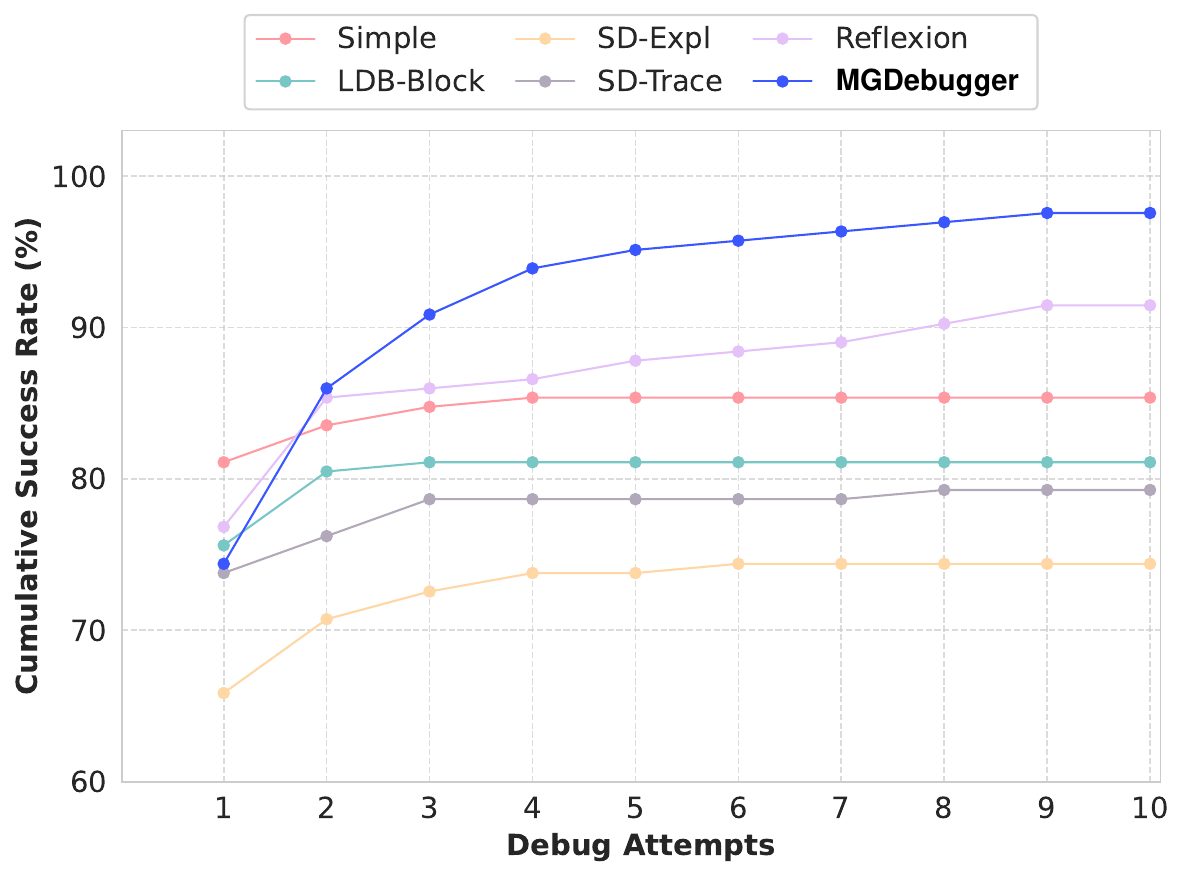}}
    \caption{Impact of debug attempts on the cumulative repair success rate of different methods.}
    \vspace{-0.25cm}
    \label{fig:debug_attempts}
\end{figure*}

Table~\ref{tab:humanevalfix} presents the RSRs across various bug categories with. We observe that \approach consistently outperforms other methods with significantly higher overall accuracies. Specifically, \approach achieves a remarkable repair success rate of 97.6\% using DeepSeek-Coder, with 100\% success rates in all bug categories except for value misuse. This is particularly notable given the complexity and diversity of the bugs in the dataset. Statistical analysis using the Wilcoxon signed-rank test with 10 repetitions confirms that these improvements are significant (p < 0.005). This highlights the effectiveness of the hierarchical debugging strategy. 

Looking into details of different bug categories, \approach shows a strong advantage in debugging bottom-level bugs, such as missing logic and excess logic. Missing logic refers to situations where essential code is omitted, preventing the solution from functioning correctly. Excess logic, on the other hand, involves unnecessary code that can lead to mistakes and confusion~\citep{muennighoff2023octopack}. 
Other methods often struggle to identify and address these underlying issues because they treat the code holistically. This can lead to confusion over bottom-level details when dealing with complex logical errors. By contrast, the hierarchical decomposition in \approach allows it to focus on different levels of code granularity. This enables more effective identification and correction of bugs. These results demonstrate the robustness and versatility of \approach across various bug types.

\subsection{Performance Consistency (RQ4)}
\label{sec:code_length}

We further assess the versatility of \approach in debugging code of varying lengths (i.e., number of tokens), since code length often correlates with complexity and debugging challenges. \edit{We categorize code snippets into short, medium, and long groups based on the 1/3 and 2/3 quantiles of code length distribution, ensuring equal sample sizes across the three categories.} We subsequently analyze the repair success rate scores obtained by \approach and baselines when using DeepSeek-Coder as the backbone LLM.

The results are presented in Figure~\ref{fig:code_length}. We observe that as the code length increases, most methods experience an obvious decrease in performance due to the increased complexity. Notably, \approach consistently outperforms other methods across different code lengths on all three datasets. This performance advantage is particularly pronounced when debugging longer and more complex code snippets on HumanEval and HumanEvalFix, where \approach maintains higher repair success rates compared to baseline methods.

\begin{table}[t]
    \centering
   \caption{Number of bugs fixed in Defects4J.}
   \resizebox{\linewidth}{!}{
   \begin{tabular}{lcccccc|c|c}
   \toprule
   \textbf{Method} & \textbf{Chart} & \textbf{Closure} & \textbf{Lang} & \textbf{Math} & \textbf{Mockito} & \textbf{Time} & \textbf{V1.2} & \textbf{V2.0} \\
   \midrule
   \textbf{CodexRepair} & 9 & 30 & 22 & 29 & 6 & 3 & 99 & 31 \\
   \textbf{GAMMA} & 11 & 24 & 16 & 25 & 3 & 3 & 82 & 45 \\
   \textbf{ThinkRepair} & 11 & 31 & 19 & 27 & 6 & 4 & 98 & 47 \\
   \textbf{ContrastRepair} & 12 & 32 & 19 & 30 & 8 & 2 & 103 & 40 \\
   \textbf{Mulpor} & 13 & 26 & 19 & 23 & 7 & 4 & 92 & \textbf{59} \\
   \textbf{ChatRepair} & \textbf{15} & 37 & 21 & 32 & 6 & 3 & 114 & 48 \\
   \textbf{\approach} & 13 & \textbf{40} & \textbf{27} & \textbf{40} & \textbf{9} & \textbf{5} & \textbf{129} & \textbf{64} \\
   \bottomrule
   \end{tabular}
   }
   \label{tab:defects4j}
\end{table}

Another important factor for LLM-based debugging is the number of debugging attempts. Iterative debugging retries allows LLMs to refine their corrections over multiple passes, potentially leading to better outcomes~\cite{yin2024thinkrepair,zhong2024debug}. We aim to assess \approach's ability to improve over successive iterations. Following~\citet{zhong2024debug}, we vary the number of debugging attempts from 1 to 10 using DeepSeek-Coder on different dataset and collect the cumulative repair success rate (RSR) scores for each method. \edit{We clarify that a single iteration refers to one complete bottom-up traversal of the function decomposition tree, including test case generation and simulated execution for all subfunctions. Even if a function is decomposed into multiple subfunctions, the process is counted as a single iteration if it involves one full pass from the leaves to the root.}

The results in Figure~\ref{fig:debug_attempts} show that \approach achieves the highest cumulative RSR score among all methods. As shown in Figure~\ref{fig:humaneval_retry_success_rate}, \approach demonstrates steady improvement across iterations on HumanEval dataset, effectively helping the model correct its own generation errors and reaching a significantly higher success rate compared to baselines. For MBPP in Figure~\ref{fig:mbpp_retry_success_rate}, \approach shows consistent gains with each attempt, highlighting its ability to systematically identify and fix bugs in LLM-generated code. Figure~\ref{fig:humanevalfix_retry_success_rate} shows that on HumanEvalFix, \approach achieves over 95\% accuracy after 5 attempts and continues to improve gradually to reach 97.6\%, though the improvement rate slows due to a few particularly challenging cases requiring specialized handling.

These results demonstrate \approach's strong potential in enhancing LLM-based code generation systems. By effectively debugging LLM-generated code through multiple iterations, \approach not only improves the immediate code quality but also shows promise in strengthening the overall reliability of AI code generation systems. This iterative refinement capability is particularly valuable as it allows LLMs to learn from and correct their own mistakes, potentially leading to more robust and dependable automated programming solutions.

\begin{figure}[t]
    \centering
    \includegraphics[width=0.6\linewidth]{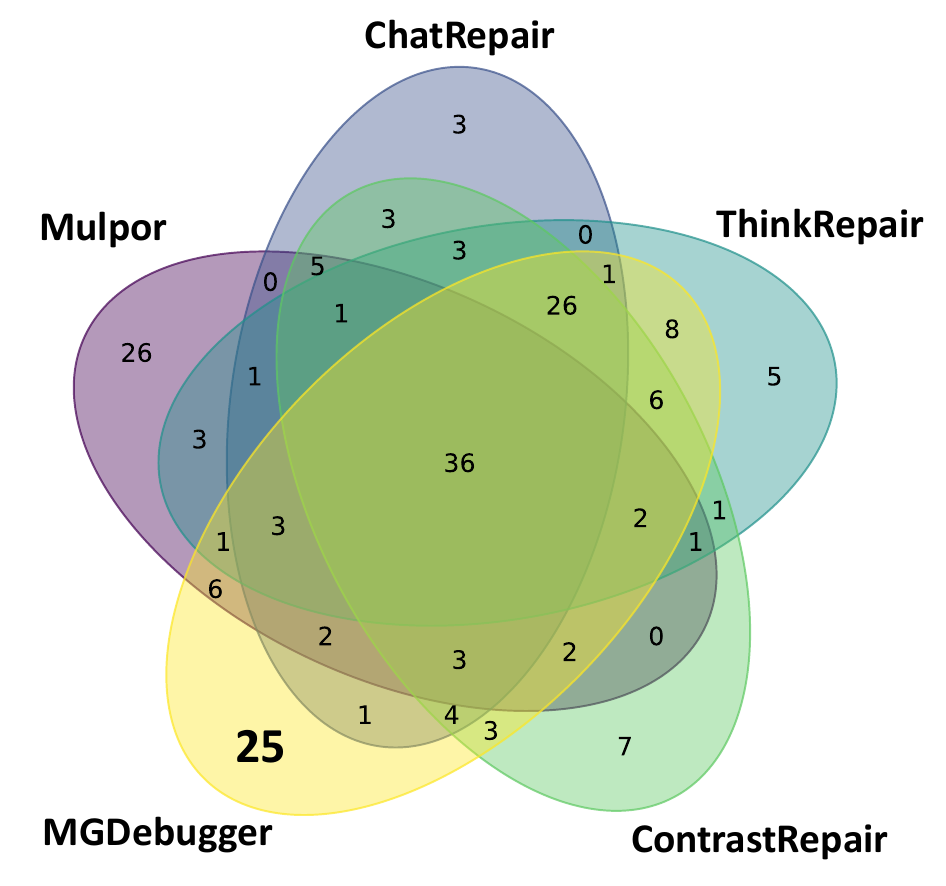}
    \vspace{-0.15cm}
    \caption{Bug fix Venn diagram in Defects4J V1.2.}
    \vspace{-0.25cm}
    \label{fig:defects4j_venn}
\end{figure}

\subsection{Generalization to Real-World Software Defects (RQ5)}\label{sec:defects4j}

While our primary focus is to explore LLMs' ability to fix LLM-generated code, we also evaluate \approach on repairing real-world, human-written bugs. We leverage the widely used Defects4J dataset~\citep{just2014defects4j}, which contains real-world defects from open-source Java projects. We compare \approach with state-of-the-art program repair approaches, including ChatRepair~\citep{xia2024automated}, Mulpor~\citep{lin2024one}, ContrastRepair~\citep{kong2025contrastrepair}, ThinkRepair~\citep{yin2024thinkrepair}, GAMMA~\citep{zhang2024gamma}, and CodexRepair~\citep{xia2023automated}.
Following prior APR tools~\citep{kong2025contrastrepair,xia2024automated,lin2024one,yin2024thinkrepair}, we use the 255 single-function bugs in Defects4J v1.2 and 82 single-line bugs in Defects4J v2.0 for evaluation and provide the perfect fault localization for each bug. To ensure a fair comparison, we use gpt-3.5-turbo-0301 as the LLM backbone and set the maximum number of generated patches to 200 following prior works~\citep{kong2025contrastrepair, xia2024automated}. \edit{For patch validation, we follow the established methodology from prior works~\citep{kong2025contrastrepair,xia2024automated,lin2024one,yin2024thinkrepair,zhang2024gamma} and conduct manual validation by three experienced engineers who independently assess each patch. A "correct" patch is defined as one that is semantically equivalent to the ground truth patch, while a "plausible" patch passes all available test cases but may not fix the underlying bug. The engineers discuss inconsistencies and reach consensus through detailed examination of patch semantics and equivalence to ground truth solutions.} 

Table~\ref{tab:defects4j} presents the number of bugs fixed by \approach and various state-of-the-art program repair approaches across different projects in Defects4J. \approach demonstrates superior overall performance, successfully fixing 129 bugs in Defects4J v1.2 and 64 bugs in Defects4J v2.0. This represents an improvement of 15 more bugs (13.2\%) over ChatRepair on Defects4J v1.2 and 5 more bugs (8.5\%) over Mulpor on Defects4J v2.0. Notably, \approach outperforms all other approaches on 5 out of 6 individual projects, with particularly strong absolute improvements.

Figure~\ref{fig:defects4j_venn} illustrates the overlap of bugs fixed by \approach and four strong baseline approaches (using ChatRepair results from the replication package of ContrastRepair~\citep{kong2025contrastrepair} due to unavailable bug id details in the original paper~\citep{xia2024automated}). The diagram shows 36 bugs fixed by all methods, while \approach uniquely fixes 25 distinct bugs, demonstrating its capability with complex defects. This complementarity with other approaches (e.g., Mulpor's 26 uniquely fixed bugs) suggests potential for combining methods to achieve even higher repair rates, as different debugging strategies appear effective for various bug types and code contexts.

Overall, these results demonstrate that \approach generalizes well beyond debugging LLM-generated code to fixing real-world software defects in production systems. The hierarchical decomposition and systematic debugging approach of \approach proves effective for human-written code with complex project structures and dependencies. This generalizability is particularly important as it suggests that \approach could serve as a unified debugging framework applicable to both AI-generated and human-written code.


\begin{figure*}[t]
    \centering
    \includegraphics[width=0.95\linewidth]{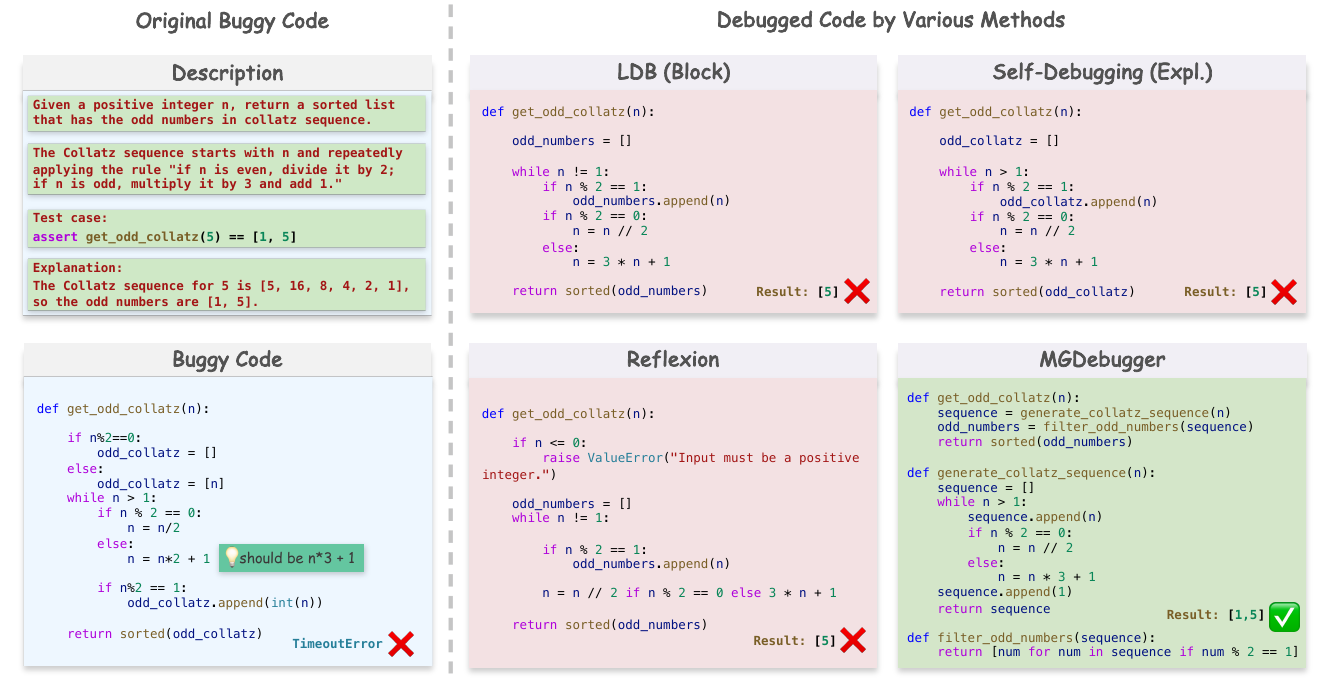}
    \vspace{-0.2cm}
    \caption{Examples of code debugging by various methods on HumanEvalFix with DeepSeek-Coder.}
    \label{fig:case_study}
    \vspace{-0.2cm}
\end{figure*}

\section{Discussion}
\subsection{Case Study}

We perform a qualitative analysis of how \approach effectively identifies and corrects buggy parts compared to baseline methods. Figure~\ref{fig:case_study} shows an example of debugging code snippets from the HumanEvalFix dataset using \approach and other representative methods, with DeepSeek-Coder-V2-Lite as the backbone LLM. The original buggy solution computes the Collatz sequence with an incorrect computation logic of $n=n\times2+1$. While other methods correct the computation to $n=n\times3+1$, they introduce a new bug that misses the last ``1" in the Collatz sequence. This is possibly because they get distracted by the need to filter odd numbers, and thus move the operation of appending the number to the results before updating $n$. \approach excelled by decomposing the problem into distinct subfunctions: sequence generation and odd number filtering. By debugging each subfunction independently, MGDebugger ensured comprehensive error correction, including the subtle requirement of incorporating 1 into the Collatz sequence. This approach demonstrates MGDebugger's ability to handle complex, multi-step problems more effectively than holistic debugging methods. Additionally, it highlights MGDebugger's ability to not only fix bugs but also restructure code for enhanced clarity and correctness, demonstrating its potential in improving the quality of LLM-generated code.




\edit{\subsection{Individual Component Analysis}
To further validate the quality of individual components, we conducted manual verification on 38 examples from HumanEval, including 29 successfully debugged and 9 failed cases, using DeepSeek-Coder-V2-Lite. Our analysis reveals three key findings: 1) \textbf{Code Decomposition}: All 38 decomposed versions are functionally equivalent to the original functions, demonstrating 100\% semantic correctness in our decomposition process. Since the overall codes have been provided, it's less likely that the model includes new bugs when transforming the code into subfunctions, demonstrating the high quality of our LLM-based decomposition approach. 2) \textbf{Test Case Generation}: 291 out of 312 generated test cases, representing 93.2\% accuracy, were correct when the model is instructed to strictly follow the public test cases when generating test cases for each subfunction. Since our approach derives test cases exclusively from the original public test cases, with a mean test coverage of 71.3\%, mutation testing scores of 9.2\% and test diversity of 28.6\%, we maintain the inherent quality characteristics of the original test suite while enhancing debugging effectiveness. Additionally, in a supplementary experiment on 9 failed debugging cases, providing additional test cases that exposed errors enabled successful debugging of 2 additional functions, indicating potential for further improvement with enhanced test diversity and coverage. 3) \textbf{Simulated Execution}: The LLM correctly predicted 389 of 413 breakpoints, achieving 94.2\% accuracy, with failures occurring only in cases requiring complex multi-step reasoning. The high accuracy aligns well with existing studies~\citep{zheng2024opencodeinterpreter,ni2024next}. Notably, when we manually corrected the prediction errors, 2 of the 9 failed debugging cases could be successfully fixed, indicating potential for further improvement with enhanced simulation accuracy. According to our manual analysis, the high quality of each component in \approach contributes to the overall effectiveness, as demonstrated by the ablation results. This suggests our approach is robust and that further improvements in component quality could lead to even better results.}

\edit{\subsection{Bug Granularity Analysis}
To validate our main claim that hierarchical debugging helps find bugs at different granularity levels, we manually analyzed the 23 problems successfully fixed by \approach using DeepSeek-Coder on HumanEval. Our analysis reveals that 13 out of 23 cases, representing 56.5\%, involved low-level bugs such as syntactic errors, API misuse, incorrect operators, and variable name errors, while 10 out of 23 cases, representing 43.5\%, involved high-level bugs including algorithmic logic errors, incorrect problem understanding, and missing core logic. 
This distribution demonstrates that \approach effectively addresses both syntactic and semantic faults across different granularity levels, validating our hierarchical debugging strategy for handling diverse error types. The ability to handle both categories showcases \approach's versatility across various debugging scenarios.}

\edit{\subsection{Evaluation on Unseen Data}

Since HumanEval and Defects4J are potentially contained in training data of LLMs, potential data leakage could affect our evaluation. To mitigate this threat, we have employed several strategies: (1) we utilized the HumanEvalFix dataset, which consists of human-crafted bugs created after the training cutoff dates of Codestral; (2) our experimental design inherently reduces the impact of potential data leakage by focusing exclusively on test cases where the No-Debugging LLM fails to produce correct solutions~\cite{olausson2023selfrepair,chen2023teaching}. If data leakage had a strong effect for a particular case, the model would likely have generated a correct solution on its first attempt, and such cases would not be included in our debugging evaluation set.

To further address potential data leakage concerns, we conducted additional experiments on LiveCodeBench~\citep{jain2025livecodebench}, which contains programming problems released after LLM training cutoffs. Using 131 problems from the February-May 2025 evaluation window with GPT-4o, \approach achieves 36.6\% accuracy with 10.8\% repair success rate, outperforming all baselines by +7.6\% over No-Debugging and +3.0\% over Reflexion (Table~\ref{tab:livecodebench}). These results also confirm that \approach's improvements are not artifacts of data leakage but represent genuine debugging capabilities on completely unseen data.

\begin{table}[t]
    \centering
    \caption{Results on LiveCodeBench with GPT-4o.}
    \vspace{-0.2cm}
    \resizebox{0.7\linewidth}{!}{
    \setlength{\tabcolsep}{3mm}{
    \begin{tabular}{lcc}
    \toprule
    \textbf{Method} & \textbf{Acc. (\%)} & \textbf{RSR (\%)} \\
    \midrule
    No-Debugging & 29.0 & -- \\
    Self-Debugging (Expl.) & 32.8 (\textcolor{cadmiumgreen}{+3.8}) & 4.3 \\
    Reflexion & 33.6 (\textcolor{cadmiumgreen}{+4.6}) & 5.4 \\
    \textbf{\approach} & \textbf{36.6} (\textcolor{cadmiumgreen}{\textbf{+7.6}}) & \textbf{10.8} \\
    \bottomrule
    \end{tabular}
    }
    }
    \label{tab:livecodebench}%
    \vspace{-0.2cm}
\end{table}}

\subsection{Threats to Validity}
While our evaluation is comprehensive, several threats to validity should be acknowledged. 1) The first is the risk of limited generalizability across programming languages and domains. To mitigate this concern, we have evaluated \approach across four distinct benchmarks spanning both Python and Java languages. This comprehensive validation across different language contexts strengthens our confidence in the approach's broad applicability. 2) Our method's performance may vary with different LLM architectures. To address this concern, we conducted experiments with various open-source LLMs of different sizes as well as several commercial models. The consistent improvements observed across this diverse set of models suggest that \approach's benefits are robust across different model architectures and capabilities. 
3) For our evaluation on Defects4J, an internal threat arises from the manual validation process used to determine the correctness of the generated patches. To mitigate this concern, we conducted thorough examinations of each patch, following established validation practices from prior work in automated program repair~\citep{xia2024automated,kong2025contrastrepair,lin2024one,yin2024thinkrepair,zhang2024gamma}.

Despite these potential limitations, our extensive evaluation across multiple dimensions—including various models, datasets, bug types, and code complexities—provides strong evidence that \approach represents a significant advancement in debugging LLM-generated codes. The consistency of our results and the magnitude of improvement over existing methods indicate broad applicability and robust performance that is likely to extend to many real-world debugging scenarios.

\section{Related Work}

\subsection{Large Language Models for Code}
Recent models such as GPT4~\citep{openai2023gpt4}, Codestral~\citep{mistralaiteam2024codestral}, and DeepSeek-Coder~\citep{zhu2024deepseekcoderv2} have advanced code generation through instruction tuning and \edit{reinforcement learning from human feedback (RLHF)} with mixed code and natural language data~\citep{ziegler2020finetuning,rafailov2023direct,anthropic2024introducing}. 
Code generation with LLMs has been enhanced by various techniques. Some approaches focus on improving the quality of generated code using planning algorithms, transitioning from outlines to detailed implementations~\citep{zhang2022planning,zelikman2023parsel,zheng2023outline}. Other methods sample multiple programs from the same LLM and rank them to identify the best one~\citep{chen2023universal,chen2022codet,ni2023lever}. Additionally, some works leverage multi-agent collaboration frameworks to enhance code generation quality~\citep{zhang2024codeagent,huang2023agentcoder,dong2024selfcollaboration}. These approaches aim to optimize the production of correct code from the outset. By contrast, \approach targets the post-generation phase, focusing on debugging and fixing errors that inevitably arise during the code generation process.

\subsection{Repairing LLM-Generated Code}
Program repair is a critical aspect of software development, aiming to automatically identify and fix bugs in code~\citep{just2014defects4j,gupta2020synthesize,yasunaga2021breakitfixit,shao2025llms}. 
There are two main streams of research in repairing code generated by LLMs: (1) training models to repair code~\citep{huang2023empirical,jiang2024training,ding2024cycle,zheng2024opencodeinterpreter,moon2024coffee,kumar2024training} and (2) providing external feedback to the raw pretrained models to fix code~\citep{jiang2023selfevolve,chen2023teaching,olausson2023selfrepair,zhong2024debug,hu2024leveraging}. By contrast to previous work that trains separate models for code repair~\citep{ding2024cycle,zheng2024opencodeinterpreter,moon2024coffee}, \approach does not require task-specific retraining but takes advantage of the inherent capabilities of pretrained LLMs. This flexibility allows \approach to operate in zero-shot settings, offering a lightweight and scalable alternative. Furthermore, exploring the ability of LLMs to fix their own code is a promising direction for self-improvement training of the LLMs~\citep{burns2023weaktostrong}.

\approach falls under the category of work that leverages pretrained models to fix code by reasoning with external feedback. Several recent methods~\citep{zhang2023selfedit,olausson2023selfrepair,bouzenia2024repairagent,lee2024unified} utilize execution results from test cases to guide LLMs in code correction. More recent works have explored advanced debugging techniques utilizing LLM's reasoning ability. 
Reflexion~\citep{shinn2023reflexion} prompts LLMs to reflect on the generated code and uses a memory buffer for iterative refinement. Self-Debugging~\citep{chen2023teaching} prompts LLMs to explain or dry run generated programs, known as rubber duck debugging. LDB~\citep{zhong2024debug} segments programs into basic blocks, tracking variable values during runtime after each block to verify the correctness against the task description. Although these methods incorporate detailed execution feedback and iterative refinement, they treat the whole function as a single unit and perform sequential debugging, limiting their effectiveness with complex code~\citep{chen2023teaching,hossain2024deep}. \approach addresses this issue by introducing a hierarchical approach, debugging from low-level errors to high-level flaws. This method ensures a more systematic and accurate debugging process, especially for complex and multifunctional systems. 

\subsection{Automated Program Repair}
Traditional APR techniques can be broadly classified into three categories: template-based~\cite{liu2019avatar,zhang2024gamma}, heuristic-based~\citep{legoues2012genprog,kim2024enhancing} and constraint based~\citep{le2017s3,mechtaev2016angelix,zhu2023tare}. While these approaches have demonstrated effectiveness, they often face challenges in scalability and patch diversity. The advent of deep learning has shifted the focus to learning-based approaches, particularly those framing program repair as a neural machine translation task~\cite{lin2024one,li2020dlfix,lutellier2020coconut,jiang2023knod}.
These methods use encoder-decoder architectures to translate buggy code into correct versions, achieving notable success but remaining limited by the quality and quantity of training data~\citep{yin2024thinkrepair}. More recently, LLM-based APR tools, powered by models like ChatGPT and Codex, have achieved superior fix rates on standard benchmarks like Defects4J~\citep{just2014defects4j} by generating diverse and contextually relevant patches~\citep{xia2023automated,xia2024automated,kong2025contrastrepair,yin2024thinkrepair,hossain2024deep,xin2024detecting,zhang2025patch}. 

\approach differs from existing APR approaches in both objective and methodology. \approach focuses on enabling LLMs to correct their own mistakes from code generation, working towards more reliable AI code generation systems. Methodologically, \approach introduces a novel bottom-up hierarchical debugging strategy combined with test case generation and LLM-simulated execution, unleashing the potential of LLMs' reasoning capabilities.

\section{Conclusion}

In this paper, we introduced \approach, a novel hierarchical code debugging framework that systematically fixes bugs at multiple levels of granularity. By decomposing complex code into a hierarchical structure, generating targeted test cases and employing LLM-simulated execution, \approach effectively identifies and fixes bugs ranging from syntax errors to logical flaws in a bottom-up manner. Experiments across various models and datasets demonstrate \approach's superior performance over existing methods, particularly in handling complex logical errors and longer code snippets.

Future work can build upon this foundation to develop more advanced code generation and debugging methodologies. One direction is to extend \approach to handle more complex bugs and code structures, such as multi-file projects and codebase with multiple dependencies. Another direction is to explore the collaboration of hierarchical code generation approaches with hierarchical debugging, enabling end-to-end code generation and debugging systems. Furthermore, integrating \approach into self-training systems to correct outputs from base models, then retraining the base models with the corrected data, could potentially improve their performance iteratively.


\subsection*{Acknowledgments}
This research is funded by the National Key Research and Development Program of China (Grant No. 2023YFB4503802) and the Natural Science Foundation of Shanghai (Grant No. 25ZR1401175). The authors would like to thank National Supercomputer Center in Guangzhou for providing high performance computational resources.

\bibliographystyle{ACM-Reference-Format}
\bibliography{mgdebugger}

@inproceedings{jain2025livecodebench,
  title={LiveCodeBench: Holistic and Contamination Free Evaluation of Large Language Models for Code},
  author={Jain, Naman and Han, King and Gu, Alex and Li, Wen-Ding and Yan, Fanjia and Zhang, Tianjun and Wang, Sida and Solar-Lezama, Armando and Sen, Koushik and Stoica, Ion},
  booktitle={The Thirteenth International Conference on Learning Representations},
  year = {2025}
}

@misc{anthropic2024introducing,
  title = {Introducing {{Claude}} 3.5 {{Sonnet}}},
  author = {{Anthropic}},
  year = {2024},
  urldate = {2025-03-14},
  howpublished = {https://www.anthropic.com/news/claude-3-5-sonnet},
  langid = {english}
}

@article{la2024code,
  title={Code simulation challenges for large language models},
  author={La Malfa, Emanuele and Weinhuber, Christoph and Torre, Orazio and Lin, Fangru and Marro, Samuele and Cohn, Anthony and Shadbolt, Nigel and Wooldridge, Michael},
  journal={arXiv preprint arXiv:2401.09074},
  year={2024}
}

@misc{austin2021program,
  title = {Program {{Synthesis}} with {{Large Language Models}}},
  author = {Austin, Jacob and Odena, Augustus and Nye, Maxwell and Bosma, Maarten and Michalewski, Henryk and Dohan, David and Jiang, Ellen and Cai, Carrie and Terry, Michael and Le, Quoc and Sutton, Charles},
  year = {2021},
  month = aug,
  number = {arXiv:2108.07732},
  eprint = {2108.07732},
  primaryclass = {cs},
  publisher = {arXiv},
  doi = {10.48550/arXiv.2108.07732},
  urldate = {2023-06-17},
  archiveprefix = {arXiv}
}

@misc{bai2023qwen,
  title = {Qwen {{Technical Report}}},
  author = {Qwen Team},
  year = {2023},
  month = sep,
  number = {arXiv:2309.16609},
  eprint = {2309.16609},
  primaryclass = {cs},
  publisher = {arXiv},
  doi = {10.48550/arXiv.2309.16609},
  urldate = {2024-09-19},
  archiveprefix = {arXiv}
}

@misc{bouzenia2024repairagent,
  title = {{{RepairAgent}}: {{An Autonomous}}, {{LLM-Based Agent}} for {{Program Repair}}},
  shorttitle = {{{RepairAgent}}},
  author = {Bouzenia, Islem and Devanbu, Premkumar and Pradel, Michael},
  year = {2024},
  month = mar,
  number = {arXiv:2403.17134},
  eprint = {2403.17134},
  primaryclass = {cs},
  publisher = {arXiv},
  doi = {10.48550/arXiv.2403.17134},
  urldate = {2024-07-30},
  archiveprefix = {arXiv}
}

@misc{burns2023weaktostrong,
  title = {Weak-to-{{Strong Generalization}}: {{Eliciting Strong Capabilities With Weak Supervision}}},
  shorttitle = {Weak-to-{{Strong Generalization}}},
  author = {Burns, Collin and Izmailov, Pavel and Kirchner, Jan Hendrik and Baker, Bowen and Gao, Leo and Aschenbrenner, Leopold and Chen, Yining and Ecoffet, Adrien and Joglekar, Manas and Leike, Jan and Sutskever, Ilya and Wu, Jeff},
  year = {2023},
  month = dec,
  number = {arXiv:2312.09390},
  eprint = {2312.09390},
  publisher = {arXiv},
  doi = {10.48550/arXiv.2312.09390},
  urldate = {2023-12-18},
  archiveprefix = {arXiv}
}

@misc{chen2021evaluating,
  title = {Evaluating {{Large Language Models Trained}} on {{Code}}},
  author = {Chen, Mark and Tworek, Jerry and Jun, Heewoo and Yuan, Qiming and Pinto, Henrique Ponde de Oliveira and Kaplan, Jared and Edwards, Harri and Burda, Yuri and Joseph, Nicholas and Brockman, Greg and Ray, Alex and Puri, Raul and Krueger, Gretchen and Petrov, Michael and Khlaaf, Heidy and Sastry, Girish and Mishkin, Pamela and Chan, Brooke and Gray, Scott and Ryder, Nick and Pavlov, Mikhail and Power, Alethea and Kaiser, Lukasz and Bavarian, Mohammad and Winter, Clemens and Tillet, Philippe and Such, Felipe Petroski and Cummings, Dave and Plappert, Matthias and Chantzis, Fotios and Barnes, Elizabeth and {Herbert-Voss}, Ariel and Guss, William Hebgen and Nichol, Alex and Paino, Alex and Tezak, Nikolas and Tang, Jie and Babuschkin, Igor and Balaji, Suchir and Jain, Shantanu and Saunders, William and Hesse, Christopher and Carr, Andrew N. and Leike, Jan and Achiam, Josh and Misra, Vedant and Morikawa, Evan and Radford, Alec and Knight, Matthew and Brundage, Miles and Murati, Mira and Mayer, Katie and Welinder, Peter and McGrew, Bob and Amodei, Dario and McCandlish, Sam and Sutskever, Ilya and Zaremba, Wojciech},
  year = {2021},
  month = jul,
  eprint = {2107.03374},
  urldate = {2022-03-30},
  archiveprefix = {arXiv}
}

@inproceedings{chen2022codet,
  title = {{{CodeT}}: {{Code Generation}} with {{Generated Tests}}},
  shorttitle = {{{CodeT}}},
  booktitle = {The {{Twelfth International Conference}} on {{Learning Representations}}},
  author = {Chen, Bei and Zhang, Fengji and Nguyen, Anh and Zan, Daoguang and Lin, Zeqi and Lou, Jian-Guang and Chen, Weizhu},
  year = {2022},
  month = nov,
  urldate = {2022-12-02}
}

@inproceedings{chen2023teaching,
  title = {Teaching {{Large Language Models}} to {{Self-Debug}}},
  booktitle = {The {{Twelfth International Conference}} on {{Learning Representations}}},
  author = {Chen, Xinyun and Lin, Maxwell and Sch{\"a}rli, Nathanael and Zhou, Denny},
  year = {2023},
  month = oct,
  urldate = {2024-07-22},
  langid = {english}
}

@misc{chen2023universal,
  title = {Universal {{Self-Consistency}} for {{Large Language Model Generation}}},
  author = {Chen, Xinyun and Aksitov, Renat and Alon, Uri and Ren, Jie and Xiao, Kefan and Yin, Pengcheng and Prakash, Sushant and Sutton, Charles and Wang, Xuezhi and Zhou, Denny},
  year = {2023},
  month = nov,
  number = {arXiv:2311.17311},
  eprint = {2311.17311},
  publisher = {arXiv},
  urldate = {2023-12-04},
  archiveprefix = {arXiv},
  langid = {english}
}

@article{ding2024cycle,
  title = {{{CYCLE}}: {{Learning}} to {{Self-Refine}} the {{Code Generation}}},
  shorttitle = {{{CYCLE}}},
  author = {Ding, Yangruibo and Min, Marcus J. and Kaiser, Gail and Ray, Baishakhi},
  year = {2024},
  month = apr,
  journal = {Proc. ACM Program. Lang.},
  volume = {8},
  number = {OOPSLA1},
  pages = {108:392--108:418},
  doi = {10.1145/3649825},
  urldate = {2024-07-25}
}

@article{dong2024selfcollaboration,
  title = {Self-Collaboration {{Code Generation}} via {{ChatGPT}}},
  author = {Dong, Yihong and Jiang, Xue and Jin, Zhi and Li, Ge},
  year = {2024},
  month = jun,
  journal = {ACM Trans. Softw. Eng. Methodol.},
  issn = {1049-331X},
  doi = {10.1145/3672459},
  urldate = {2023-12-07},
  langid = {english}
}

@misc{dou2024what,
  title = {What's {{Wrong}} with {{Your Code Generated}} by {{Large Language Models}}? {{An Extensive Study}}},
  shorttitle = {What's {{Wrong}} with {{Your Code Generated}} by {{Large Language Models}}?},
  author = {Dou, Shihan and Jia, Haoxiang and Wu, Shenxi and Zheng, Huiyuan and Zhou, Weikang and Wu, Muling and Chai, Mingxu and Fan, Jessica and Huang, Caishuang and Tao, Yunbo and Liu, Yan and Zhou, Enyu and Zhang, Ming and Zhou, Yuhao and Wu, Yueming and Zheng, Rui and Wen, Ming and Weng, Rongxiang and Wang, Jingang and Cai, Xunliang and Gui, Tao and Qiu, Xipeng and Zhang, Qi and Huang, Xuanjing},
  year = {2024},
  month = jul,
  number = {arXiv:2407.06153},
  eprint = {2407.06153},
  publisher = {arXiv},
  urldate = {2024-07-12},
  archiveprefix = {arXiv},
  langid = {english}
}

@inproceedings{gupta2020synthesize,
  title = {Synthesize, {{Execute}} and {{Debug}}: {{Learning}} to {{Repair}} for {{Neural Program Synthesis}}},
  shorttitle = {Synthesize, {{Execute}} and {{Debug}}},
  booktitle = {Advances in {{Neural Information Processing Systems}}},
  author = {Gupta, Kavi and Christensen, Peter Ebert and Chen, Xinyun and Song, Dawn},
  year = {2020},
  volume = {33},
  pages = {17685--17695},
  publisher = {Curran Associates, Inc.},
  urldate = {2024-09-10}
}

@inproceedings{hendrycks2021measuring,
  title = {Measuring {{Coding Challenge Competence With APPS}}},
  booktitle = {Thirty-Fifth {{Conference}} on {{Neural Information Processing Systems Datasets}} and {{Benchmarks Track}} ({{Round}} 2)},
  author = {Hendrycks, Dan and Basart, Steven and Kadavath, Saurav and Mazeika, Mantas and Arora, Akul and Guo, Ethan and Burns, Collin and Puranik, Samir and He, Horace and Song, Dawn and Steinhardt, Jacob},
  year = {2021},
  month = aug,
  urldate = {2024-02-27},
  langid = {english}
}

@article{hossain2024deep,
  title = {A {{Deep Dive}} into {{Large Language Models}} for {{Automated Bug Localization}} and {{Repair}}},
  author = {Hossain, Soneya Binta and Jiang, Nan and Zhou, Qiang and Li, Xiaopeng and Chiang, Wen-Hao and Lyu, Yingjun and Nguyen, Hoan and Tripp, Omer},
  year = {2024},
  month = jul,
  journal = {Proc. ACM Softw. Eng.},
  volume = {1},
  number = {FSE},
  pages = {66:1471--66:1493},
  doi = {10.1145/3660773},
  urldate = {2024-09-16}
}

@misc{hu2024leveraging,
  title = {Leveraging {{Print Debugging}} to {{Improve Code Generation}} in {{Large Language Models}}},
  author = {Hu, Xueyu and Kuang, Kun and Sun, Jiankai and Yang, Hongxia and Wu, Fei},
  year = {2024},
  month = jan,
  number = {arXiv:2401.05319},
  eprint = {2401.05319},
  publisher = {arXiv},
  urldate = {2024-02-27},
  archiveprefix = {arXiv},
  langid = {english}
}

@misc{huang2023agentcoder,
  title = {{{AgentCoder}}: {{Multi-Agent-based Code Generation}} with {{Iterative Testing}} and {{Optimisation}}},
  shorttitle = {{{AgentCoder}}},
  author = {Huang, Dong and Bu, Qingwen and Zhang, Jie M. and Luck, Michael and Cui, Heming},
  year = {2023},
  month = dec,
  number = {arXiv:2312.13010},
  eprint = {2312.13010},
  publisher = {arXiv},
  doi = {10.48550/arXiv.2312.13010},
  urldate = {2023-12-23},
  archiveprefix = {arXiv}
}

@inproceedings{huang2023empirical,
  title = {An {{Empirical Study}} on {{Fine-Tuning Large Language Models}} of {{Code}} for {{Automated Program Repair}}},
  booktitle = {2023 38th {{IEEE}}/{{ACM International Conference}} on {{Automated Software Engineering}} ({{ASE}})},
  author = {Huang, Kai and Meng, Xiangxin and Zhang, Jian and Liu, Yang and Wang, Wenjie and Li, Shuhao and Zhang, Yuqing},
  year = {2023},
  month = sep,
  pages = {1162--1174},
  publisher = {IEEE},
  address = {Luxembourg, Luxembourg},
  doi = {10.1109/ASE56229.2023.00181},
  urldate = {2024-09-16},
  isbn = {979-8-3503-2996-4}
}

@book{isazadeh2017source,
  title = {Source {{Code Modularization}}},
  author = {Isazadeh, Ayaz and Izadkhah, Habib and Elgedawy, Islam},
  year = {2017},
  publisher = {Springer International Publishing},
  address = {Cham},
  doi = {10.1007/978-3-319-63346-6},
  urldate = {2024-09-15},
  isbn = {978-3-319-63344-2 978-3-319-63346-6},
  langid = {english}
}

@misc{jain2023llmassisted,
  title = {{{LLM-Assisted Code Cleaning For Training Accurate Code Generators}}},
  author = {Jain, Naman and Zhang, Tianjun and Chiang, Wei-Lin and Gonzalez, Joseph E. and Sen, Koushik and Stoica, Ion},
  year = {2023},
  month = nov,
  number = {arXiv:2311.14904},
  eprint = {2311.14904},
  publisher = {arXiv},
  doi = {10.48550/arXiv.2311.14904},
  urldate = {2023-12-11},
  archiveprefix = {arXiv}
}

@inproceedings{jiang2023knod,
  title = {{{KNOD}}: {{Domain Knowledge Distilled Tree Decoder}} for {{Automated Program Repair}}},
  shorttitle = {{{KNOD}}},
  booktitle = {Proceedings of the 45th {{International Conference}} on {{Software Engineering}}},
  author = {Jiang, Nan and Lutellier, Thibaud and Lou, Yiling and Tan, Lin and Goldwasser, Dan and Zhang, Xiangyu},
  year = {2023},
  month = jul,
  series = {{{ICSE}} '23},
  pages = {1251--1263},
  publisher = {IEEE Press},
  address = {Melbourne, Victoria, Australia},
  doi = {10.1109/ICSE48619.2023.00111},
  urldate = {2024-11-14},
  isbn = {978-1-6654-5701-9}
}

@misc{jiang2023selfevolve,
  title = {{{SelfEvolve}}: {{A Code Evolution Framework}} via {{Large Language Models}}},
  shorttitle = {{{SelfEvolve}}},
  author = {Jiang, Shuyang and Wang, Yuhao and Wang, Yu},
  year = {2023},
  month = jun,
  number = {arXiv:2306.02907},
  eprint = {2306.02907},
  primaryclass = {cs},
  publisher = {arXiv},
  doi = {10.48550/arXiv.2306.02907},
  urldate = {2024-09-09},
  archiveprefix = {arXiv}
}

@inproceedings{jiang2024training,
  title = {Training {{LLMs}} to {{Better Self-Debug}} and {{Explain Code}}},
  booktitle = {{{NeurIPS2024}}},
  author = {Jiang, Nan and Li, Xiaopeng and Wang, Shiqi and Zhou, Qiang and Hossain, Soneya Binta and Ray, Baishakhi and Kumar, Varun and Ma, Xiaofei and Deoras, Anoop},
  year = {2024},
  month = may,
  urldate = {2024-09-09},
  langid = {english}
}

@inproceedings{just2014defects4j,
  title = {{{Defects4J}}: A Database of Existing Faults to Enable Controlled Testing Studies for {{Java}} Programs},
  shorttitle = {{{Defects4J}}},
  booktitle = {Proceedings of the 2014 {{International Symposium}} on {{Software Testing}} and {{Analysis}}},
  author = {Just, Ren{\'e} and Jalali, Darioush and Ernst, Michael D.},
  year = {2014},
  month = jul,
  series = {{{ISSTA}} 2014},
  pages = {437--440},
  publisher = {Association for Computing Machinery},
  address = {New York, NY, USA},
  doi = {10.1145/2610384.2628055},
  urldate = {2024-07-24},
  isbn = {978-1-4503-2645-2}
}

@inproceedings{kim2024enhancing,
  title = {Enhancing the {{Efficiency}} of {{Automated Program Repair}} via {{Greybox Analysis}}},
  booktitle = {Proceedings of the 39th {{IEEE}}/{{ACM International Conference}} on {{Automated Software Engineering}}},
  author = {Kim, YoungJae and Park, Yechan and Han, Seungheon and Yi, Jooyong},
  year = {2024},
  month = oct,
  series = {{{ASE}} '24},
  pages = {1719--1731},
  publisher = {Association for Computing Machinery},
  address = {New York, NY, USA},
  doi = {10.1145/3691620.3695602},
  urldate = {2025-03-15},
  isbn = {979-8-4007-1248-7}
}

@article{kong2025contrastrepair,
  title = {{{ContrastRepair}}: {{Enhancing Conversation-Based Automated Program Repair}} via {{Contrastive Test Case Pairs}}},
  shorttitle = {{{ContrastRepair}}},
  author = {Kong, Jiaolong and Xie, Xiaofei and Cheng, Mingfei and Liu, Shangqing and Du, Xiaoning and Guo, Qi},
  year = {2025},
  month = mar,
  journal = {ACM Trans. Softw. Eng. Methodol.},
  issn = {1049-331X},
  doi = {10.1145/3719345},
  urldate = {2025-03-14}
}

@misc{kumar2024training,
  title = {Training {{Language Models}} to {{Self-Correct}} via {{Reinforcement Learning}}},
  author = {Kumar, Aviral and Zhuang, Vincent and Agarwal, Rishabh and Su, Yi and {Co-Reyes}, John D. and Singh, Avi and Baumli, Kate and Iqbal, Shariq and Bishop, Colton and Roelofs, Rebecca and Zhang, Lei M. and McKinney, Kay and Shrivastava, Disha and Paduraru, Cosmin and Tucker, George and Precup, Doina and Behbahani, Feryal and Faust, Aleksandra},
  year = {2024},
  month = sep,
  number = {arXiv:2409.12917},
  eprint = {2409.12917},
  primaryclass = {cs},
  publisher = {arXiv},
  urldate = {2024-09-22},
  archiveprefix = {arXiv},
  langid = {english}
}

@inproceedings{le2017s3,
  title = {S3: Syntax- and Semantic-Guided Repair Synthesis via Programming by Examples},
  shorttitle = {S3},
  booktitle = {Proceedings of the 2017 11th {{Joint Meeting}} on {{Foundations}} of {{Software Engineering}}},
  author = {Le, Xuan-Bach D. and Chu, Duc-Hiep and Lo, David and Le Goues, Claire and Visser, Willem},
  year = {2017},
  month = aug,
  series = {{{ESEC}}/{{FSE}} 2017},
  pages = {593--604},
  publisher = {Association for Computing Machinery},
  address = {New York, NY, USA},
  doi = {10.1145/3106237.3106309},
  urldate = {2025-03-15},
  isbn = {978-1-4503-5105-8}
}

@misc{lee2024unified,
  title = {A {{Unified Debugging Approach}} via {{LLM-Based Multi-Agent Synergy}}},
  author = {Lee, Cheryl and Xia, Chunqiu Steven and Huang, Jen-tse and Zhu, Zhouruixin and Zhang, Lingming and Lyu, Michael R.},
  year = {2024},
  month = apr,
  number = {arXiv:2404.17153},
  eprint = {2404.17153},
  publisher = {arXiv},
  urldate = {2024-07-30},
  archiveprefix = {arXiv},
  langid = {english}
}

@article{legoues2012genprog,
  title = {{{GenProg}}: {{A Generic Method}} for {{Automatic Software Repair}}},
  shorttitle = {{{GenProg}}},
  author = {Le Goues, Claire and Nguyen, ThanhVu and Forrest, Stephanie and Weimer, Westley},
  year = {2012},
  month = jan,
  journal = {IEEE Trans. Softw. Eng.},
  volume = {38},
  number = {1},
  pages = {54--72},
  issn = {0098-5589},
  doi = {10.1109/TSE.2011.104},
  urldate = {2025-03-15}
}

@inproceedings{li2020dlfix,
  title = {{{DLFix}}: Context-Based Code Transformation Learning for Automated Program Repair},
  shorttitle = {{{DLFix}}},
  booktitle = {Proceedings of the {{ACM}}/{{IEEE}} 42nd {{International Conference}} on {{Software Engineering}}},
  author = {Li, Yi and Wang, Shaohua and Nguyen, Tien N.},
  year = {2020},
  month = oct,
  series = {{{ICSE}} '20},
  pages = {602--614},
  publisher = {Association for Computing Machinery},
  address = {New York, NY, USA},
  doi = {10.1145/3377811.3380345},
  urldate = {2025-03-15},
  isbn = {978-1-4503-7121-6}
}

@article{li2022competitionlevel,
  title = {Competition-Level Code Generation with {{AlphaCode}}},
  author = {Li, Yujia and Choi, David and Chung, Junyoung and Kushman, Nate and Schrittwieser, Julian and Leblond, R{\'e}mi and Eccles, Tom and Keeling, James and Gimeno, Felix and Dal Lago, Agustin and Hubert, Thomas and Choy, Peter and {de Masson d'Autume}, Cyprien and Babuschkin, Igor and Chen, Xinyun and Huang, Po-Sen and Welbl, Johannes and Gowal, Sven and Cherepanov, Alexey and Molloy, James and Mankowitz, Daniel J. and Sutherland Robson, Esme and Kohli, Pushmeet and {de Freitas}, Nando and Kavukcuoglu, Koray and Vinyals, Oriol},
  year = {2022},
  month = dec,
  journal = {Science},
  volume = {378},
  number = {6624},
  pages = {1092--1097},
  publisher = {American Association for the Advancement of Science},
  doi = {10.1126/science.abq1158},
  urldate = {2024-09-18}
}

@misc{li2023chain,
  title = {Chain of {{Code}}: {{Reasoning}} with a {{Language Model-Augmented Code Emulator}}},
  shorttitle = {Chain of {{Code}}},
  author = {Li, Chengshu and Liang, Jacky and Zeng, Andy and Chen, Xinyun and Hausman, Karol and Sadigh, Dorsa and Levine, Sergey and {Fei-Fei}, Li and Xia, Fei and Ichter, Brian},
  year = {2023},
  month = dec,
  number = {arXiv:2312.04474},
  eprint = {2312.04474},
  publisher = {arXiv},
  urldate = {2023-12-08},
  archiveprefix = {arXiv},
  langid = {english}
}

@inproceedings{lin2024one,
  title = {One {{Size Does Not Fit All}}: {{Multi-granularity Patch Generation}} for {{Better Automated Program Repair}}},
  shorttitle = {One {{Size Does Not Fit All}}},
  booktitle = {Proceedings of the 33rd {{ACM SIGSOFT International Symposium}} on {{Software Testing}} and {{Analysis}}},
  author = {Lin, Bo and Wang, Shangwen and Wen, Ming and Chen, Liqian and Mao, Xiaoguang},
  year = {2024},
  month = sep,
  series = {{{ISSTA}} 2024},
  pages = {1554--1566},
  publisher = {Association for Computing Machinery},
  address = {New York, NY, USA},
  doi = {10.1145/3650212.3680381},
  urldate = {2025-01-20},
  isbn = {979-8-4007-0612-7}
}

@inproceedings{liu2019avatar,
  title = {{{AVATAR}}: {{Fixing Semantic Bugs}} with {{Fix Patterns}} of {{Static Analysis Violations}}},
  shorttitle = {{{AVATAR}}},
  booktitle = {2019 {{IEEE}} 26th {{International Conference}} on {{Software Analysis}}, {{Evolution}} and {{Reengineering}} ({{SANER}})},
  author = {Liu, Kui and Koyuncu, Anil and Kim, Dongsun and Bissyand{\`e}, Tegawende F.},
  year = {2019},
  month = feb,
  pages = {1--12},
  publisher = {IEEE Computer Society},
  issn = {1534-5351},
  doi = {10.1109/SANER.2019.8667970},
  urldate = {2025-03-15},
  isbn = {978-1-7281-0591-8},
  langid = {english}
}

@article{liu2023rltf,
  title = {{{RLTF}}: {{Reinforcement Learning}} from {{Unit Test Feedback}}},
  shorttitle = {{{RLTF}}},
  author = {Liu, Jiate and Zhu, Yiqin and Xiao, Kaiwen and Fu, Qiang and Han, Xiao and Wei, Yang and Ye, Deheng},
  year = {2023},
  month = jul,
  journal = {Transactions on Machine Learning Research},
  issn = {2835-8856},
  urldate = {2024-02-27},
  langid = {english}
}

@misc{liu2023your,
  title = {Is {{Your Code Generated}} by {{ChatGPT Really Correct}}? {{Rigorous Evaluation}} of {{Large Language Models}} for {{Code Generation}}},
  shorttitle = {Is {{Your Code Generated}} by {{ChatGPT Really Correct}}?},
  author = {Liu, Jiawei and Xia, Chunqiu Steven and Wang, Yuyao and Zhang, Lingming},
  year = {2023},
  month = oct,
  number = {arXiv:2305.01210},
  eprint = {2305.01210},
  primaryclass = {cs},
  publisher = {arXiv},
  doi = {10.48550/arXiv.2305.01210},
  urldate = {2023-11-01},
  archiveprefix = {arXiv}
}

@inproceedings{lutellier2020coconut,
  title = {{{CoCoNuT}}: Combining Context-Aware Neural Translation Models Using Ensemble for Program Repair},
  shorttitle = {{{CoCoNuT}}},
  booktitle = {Proceedings of the 29th {{ACM SIGSOFT International Symposium}} on {{Software Testing}} and {{Analysis}}},
  author = {Lutellier, Thibaud and Pham, Hung Viet and Pang, Lawrence and Li, Yitong and Wei, Moshi and Tan, Lin},
  year = {2020},
  month = jul,
  series = {{{ISSTA}} 2020},
  pages = {101--114},
  publisher = {Association for Computing Machinery},
  address = {New York, NY, USA},
  doi = {10.1145/3395363.3397369},
  urldate = {2025-03-15},
  isbn = {978-1-4503-8008-9}
}

@inproceedings{mechtaev2016angelix,
  title = {Angelix: Scalable Multiline Program Patch Synthesis via Symbolic Analysis},
  shorttitle = {Angelix},
  booktitle = {Proceedings of the 38th {{International Conference}} on {{Software Engineering}}},
  author = {Mechtaev, Sergey and Yi, Jooyong and Roychoudhury, Abhik},
  year = {2016},
  month = may,
  series = {{{ICSE}} '16},
  pages = {691--701},
  publisher = {Association for Computing Machinery},
  address = {New York, NY, USA},
  doi = {10.1145/2884781.2884807},
  urldate = {2025-03-15},
  isbn = {978-1-4503-3900-1}
}

@misc{mistralaiteam2024codestral,
  title = {Codestral: {{Hello}}, {{World}}!},
  shorttitle = {Codestral},
  author = {{Mistral AI team}},
  year = {2024},
  month = may,
  urldate = {2024-09-09},
  chapter = {news},
  howpublished = {https://mistral.ai/news/codestral/},
  langid = {american}
}

@misc{moon2024coffee,
  title = {Coffee: {{Boost Your Code LLMs}} by {{Fixing Bugs}} with {{Feedback}}},
  shorttitle = {Coffee},
  author = {Moon, Seungjun and Chae, Hyungjoo and Song, Yongho and Kwon, Taeyoon and Kang, Dongjin and Ong, Kai Tzu-iunn and Hwang, Seung-won and Yeo, Jinyoung},
  year = {2024},
  month = feb,
  number = {arXiv:2311.07215},
  eprint = {2311.07215},
  primaryclass = {cs},
  publisher = {arXiv},
  urldate = {2024-09-10},
  archiveprefix = {arXiv},
  langid = {english}
}

@inproceedings{muennighoff2023octopack,
  title = {{{OctoPack}}: {{Instruction Tuning Code Large Language Models}}},
  shorttitle = {{{OctoPack}}},
  booktitle = {The {{Twelfth International Conference}} on {{Learning Representations}}},
  author = {Muennighoff, Niklas and Liu, Qian and Zebaze, Armel Randy and Zheng, Qinkai and Hui, Binyuan and Zhuo, Terry Yue and Singh, Swayam and Tang, Xiangru and Werra, Leandro Von and Longpre, Shayne},
  year = {2023},
  month = oct,
  urldate = {2024-09-18},
  langid = {english}
}

@misc{ni2023lever,
  title = {{{LEVER}}: {{Learning}} to {{Verify Language-to-Code Generation}} with {{Execution}}},
  shorttitle = {{{LEVER}}},
  author = {Ni, Ansong and Iyer, Srini and Radev, Dragomir and Stoyanov, Ves and Yih, Wen-tau and Wang, Sida I. and Lin, Xi Victoria},
  year = {2023},
  month = feb,
  number = {arXiv:2302.08468},
  eprint = {2302.08468},
  publisher = {arXiv},
  doi = {10.48550/arXiv.2302.08468},
  urldate = {2023-06-17},
  archiveprefix = {arXiv}
}

@misc{ni2024next,
  title = {{{NExT}}: {{Teaching Large Language Models}} to {{Reason}} about {{Code Execution}}},
  shorttitle = {{{NExT}}},
  author = {Ni, Ansong and Allamanis, Miltiadis and Cohan, Arman and Deng, Yinlin and Shi, Kensen and Sutton, Charles and Yin, Pengcheng},
  year = {2024},
  month = apr,
  number = {arXiv:2404.14662},
  eprint = {2404.14662},
  publisher = {arXiv},
  urldate = {2024-05-06},
  archiveprefix = {arXiv},
  langid = {english}
}

@inproceedings{olausson2023selfrepair,
  title = {Is {{Self-Repair}} a {{Silver Bullet}} for {{Code Generation}}?},
  booktitle = {The {{Twelfth International Conference}} on {{Learning Representations}}},
  author = {Olausson, Theo X. and Inala, Jeevana Priya and Wang, Chenglong and Gao, Jianfeng and {Solar-Lezama}, Armando},
  year = {2023},
  month = oct,
  urldate = {2024-07-22},
  langid = {english}
}

@misc{openai2023gpt4,
  title = {{{GPT-4 Technical Report}}},
  author = {OpenAI},
  year = {2023},
  month = mar,
  number = {arXiv:2303.08774},
  eprint = {2303.08774},
  primaryclass = {cs},
  publisher = {arXiv},
  doi = {10.48550/arXiv.2303.08774},
  urldate = {2023-11-02},
  archiveprefix = {arXiv}
}

@inproceedings{rafailov2023direct,
  title = {Direct {{Preference Optimization}}: {{Your Language Model}} Is {{Secretly}} a {{Reward Model}}},
  shorttitle = {Direct {{Preference Optimization}}},
  booktitle = {Thirty-Seventh {{Conference}} on {{Neural Information Processing Systems}}},
  author = {Rafailov, Rafael and Sharma, Archit and Mitchell, Eric and Manning, Christopher D. and Ermon, Stefano and Finn, Chelsea},
  year = {2023},
  month = nov,
  urldate = {2024-01-08},
  langid = {english}
}

@article{schafer2024empirical,
  title = {An {{Empirical Evaluation}} of {{Using Large Language Models}} for {{Automated Unit Test Generation}}},
  author = {Sch{\"a}fer, Max and Nadi, Sarah and Eghbali, Aryaz and Tip, Frank},
  year = {2024},
  month = jan,
  journal = {IEEE Transactions on Software Engineering},
  volume = {50},
  number = {1},
  pages = {85--105},
  issn = {1939-3520},
  doi = {10.1109/TSE.2023.3334955},
  urldate = {2024-05-11}
}

@inproceedings{shinn2023reflexion,
  title = {Reflexion: Language Agents with Verbal Reinforcement Learning},
  shorttitle = {Reflexion},
  booktitle = {Thirty-Seventh {{Conference}} on {{Neural Information Processing Systems}}},
  author = {Shinn, Noah and Cassano, Federico and Gopinath, Ashwin and Narasimhan, Karthik R. and Yao, Shunyu},
  year = {2023},
  month = nov,
  urldate = {2023-12-26},
  langid = {english}
}

@misc{tian2024debugbench,
  title = {{{DebugBench}}: {{Evaluating Debugging Capability}} of {{Large Language Models}}},
  shorttitle = {{{DebugBench}}},
  author = {Tian, Runchu and Ye, Yining and Qin, Yujia and Cong, Xin and Lin, Yankai and Pan, Yinxu and Wu, Yesai and Liu, Zhiyuan and Sun, Maosong},
  year = {2024},
  month = jan,
  number = {arXiv:2401.04621},
  eprint = {2401.04621},
  publisher = {arXiv},
  doi = {10.48550/arXiv.2401.04621},
  urldate = {2024-05-13},
  archiveprefix = {arXiv}
}

@misc{touvron2023llama,
  title = {{{LLaMA}}: {{Open}} and {{Efficient Foundation Language Models}}},
  shorttitle = {{{LLaMA}}},
  author = {Touvron, Hugo and Lavril, Thibaut and Izacard, Gautier and Martinet, Xavier and Lachaux, Marie-Anne and Lacroix, Timoth{\'e}e and Rozi{\`e}re, Baptiste and Goyal, Naman and Hambro, Eric and Azhar, Faisal and Rodriguez, Aurelien and Joulin, Armand and Grave, Edouard and Lample, Guillaume},
  year = {2023},
  month = feb,
  number = {arXiv:2302.13971},
  eprint = {2302.13971},
  primaryclass = {cs},
  publisher = {arXiv},
  doi = {10.48550/arXiv.2302.13971},
  urldate = {2023-06-17},
  archiveprefix = {arXiv}
}

@inproceedings{wang2021automatic,
  title = {Automatic {{Unit Test Generation}} for {{Machine Learning Libraries}}: {{How Far Are We}}?},
  shorttitle = {Automatic {{Unit Test Generation}} for {{Machine Learning Libraries}}},
  booktitle = {2021 {{IEEE}}/{{ACM}} 43rd {{International Conference}} on {{Software Engineering}} ({{ICSE}})},
  author = {Wang, Song and Shrestha, Nishtha and Subburaman, Abarna Kucheri and Wang, Junjie and Wei, Moshi and Nagappan, Nachiappan},
  year = {2021},
  month = may,
  pages = {1548--1560},
  issn = {1558-1225},
  doi = {10.1109/ICSE43902.2021.00138},
  urldate = {2024-05-11}
}

@inproceedings{wen2024learning,
  title = {Learning {{Task Decomposition}} to {{Assist Humans}} in {{Competitive Programming}}},
  booktitle = {Proceedings of the 62nd {{Annual Meeting}} of the {{Association}} for {{Computational Linguistics}} ({{Volume}} 1: {{Long Papers}})},
  author = {Wen, Jiaxin and Zhong, Ruiqi and Ke, Pei and Shao, Zhihong and Wang, Hongning and Huang, Minlie},
  editor = {Ku, Lun-Wei and Martins, Andre and Srikumar, Vivek},
  year = {2024},
  month = aug,
  pages = {11700--11723},
  publisher = {Association for Computational Linguistics},
  address = {Bangkok, Thailand},
  doi = {10.18653/v1/2024.acl-long.629},
  urldate = {2024-10-06}
}

@inproceedings{woodfield1981effect,
  title = {The Effect of Modularization and Comments on Program Comprehension},
  booktitle = {Proceedings of the 5th International Conference on {{Software}} Engineering},
  author = {Woodfield, S. N. and Dunsmore, H. E. and Shen, V. Y.},
  year = {1981},
  month = mar,
  series = {{{ICSE}} '81},
  pages = {215--223},
  publisher = {IEEE Press},
  address = {San Diego, California, USA},
  urldate = {2024-09-14},
  isbn = {978-0-89791-146-7}
}

@inproceedings{xia2023automated,
  title = {Automated {{Program Repair}} in the {{Era}} of {{Large Pre-trained Language Models}}},
  booktitle = {2023 {{IEEE}}/{{ACM}} 45th {{International Conference}} on {{Software Engineering}} ({{ICSE}})},
  author = {Xia, Chunqiu Steven and Wei, Yuxiang and Zhang, Lingming},
  year = {2023},
  month = may,
  pages = {1482--1494},
  issn = {1558-1225},
  doi = {10.1109/ICSE48619.2023.00129},
  urldate = {2024-05-13}
}

@inproceedings{xia2024automated,
  title = {Automated {{Program Repair}} via {{Conversation}}: {{Fixing}} 162 out of 337 {{Bugs}} for \$0.42 {{Each}} Using {{ChatGPT}}},
  shorttitle = {Automated {{Program Repair}} via {{Conversation}}},
  booktitle = {Proceedings of the 33rd {{ACM SIGSOFT International Symposium}} on {{Software Testing}} and {{Analysis}}},
  author = {Xia, Chunqiu Steven and Zhang, Lingming},
  year = {2024},
  month = sep,
  pages = {819--831},
  publisher = {ACM},
  address = {Vienna Austria},
  doi = {10.1145/3650212.3680323},
  urldate = {2025-01-13},
  isbn = {979-8-4007-0612-7},
  langid = {english}
}

@article{xin2024detecting,
  title = {Detecting, {{Creating}}, {{Repairing}}, and {{Understanding Indivisible Multi-Hunk Bugs}}},
  author = {Xin, Qi and Wu, Haojun and Tang, Jinran and Liu, Xinyu and Reiss, Steven P. and Xuan, Jifeng},
  year = {2024},
  month = jul,
  journal = {Reproduction Package for Article Detecting, Creating, Repairing, and Understanding Indivisible Multi-Hunk Bugs},
  volume = {1},
  number = {FSE},
  pages = {121:2747--121:2770},
  doi = {10.1145/3660828},
  urldate = {2025-03-15}
}

@inproceedings{yasunaga2021breakitfixit,
  title = {Break-{{It-Fix-It}}: {{Unsupervised Learning}} for {{Program Repair}}},
  shorttitle = {Break-{{It-Fix-It}}},
  booktitle = {Proceedings of the 38th {{International Conference}} on {{Machine Learning}}},
  author = {Yasunaga, Michihiro and Liang, Percy},
  year = {2021},
  month = jul,
  pages = {11941--11952},
  publisher = {PMLR},
  issn = {2640-3498},
  urldate = {2024-07-25},
  langid = {english}
}

@inproceedings{yin2024thinkrepair,
  title = {{{ThinkRepair}}: {{Self-Directed Automated Program Repair}}},
  shorttitle = {{{ThinkRepair}}},
  booktitle = {Proceedings of the 33rd {{ACM SIGSOFT International Symposium}} on {{Software Testing}} and {{Analysis}}},
  author = {Yin, Xin and Ni, Chao and Wang, Shaohua and Li, Zhenhao and Zeng, Limin and Yang, Xiaohu},
  year = {2024},
  month = sep,
  series = {{{ISSTA}} 2024},
  pages = {1274--1286},
  publisher = {Association for Computing Machinery},
  address = {New York, NY, USA},
  doi = {10.1145/3650212.3680359},
  urldate = {2024-09-17},
  isbn = {979-8-4007-0612-7}
}

@inproceedings{zelikman2023parsel,
  title = {Parsel: {{Algorithmic Reasoning}} with {{Language Models}} by {{Composing Decompositions}}},
  shorttitle = {Parsel},
  booktitle = {Thirty-Seventh {{Conference}} on {{Neural Information Processing Systems}}},
  author = {Zelikman, Eric and Huang, Qian and Poesia, Gabriel and Goodman, Noah and Haber, Nick},
  year = {2023},
  month = nov,
  urldate = {2023-12-26},
  langid = {english}
}

@book{zeller2009why,
  title = {Why Programs Fail: A Guide to Systematic Debugging},
  shorttitle = {Why Programs Fail},
  author = {Zeller, Andreas},
  year = {2009},
  publisher = {Morgan Kaufmann},
  urldate = {2024-09-13}
}

@inproceedings{zhang2022planning,
  title = {Planning with {{Large Language Models}} for {{Code Generation}}},
  booktitle = {The {{Eleventh International Conference}} on {{Learning Representations}}},
  author = {Zhang, Shun and Chen, Zhenfang and Shen, Yikang and Ding, Mingyu and Tenenbaum, Joshua B. and Gan, Chuang},
  year = {2022},
  month = sep,
  urldate = {2023-11-30},
  langid = {english}
}

@inproceedings{zhang2023selfedit,
  title = {Self-{{Edit}}: {{Fault-Aware Code Editor}} for {{Code Generation}}},
  shorttitle = {Self-{{Edit}}},
  booktitle = {Proceedings of the 61st {{Annual Meeting}} of the {{Association}} for {{Computational Linguistics}} ({{Volume}} 1: {{Long Papers}})},
  author = {Zhang, Kechi and Li, Zhuo and Li, Jia and Li, Ge and Jin, Zhi},
  year = {2023},
  month = jul,
  pages = {769--787},
  publisher = {Association for Computational Linguistics},
  address = {Toronto, Canada},
  doi = {10.18653/v1/2023.acl-long.45},
  urldate = {2024-07-25}
}

@misc{zhang2024codeagent,
  title = {{{CodeAgent}}: {{Enhancing Code Generation}} with {{Tool-Integrated Agent Systems}} for {{Real-World Repo-level Coding Challenges}}},
  shorttitle = {{{CodeAgent}}},
  author = {Zhang, Kechi and Li, Jia and Li, Ge and Shi, Xianjie and Jin, Zhi},
  year = {2024},
  month = jan,
  number = {arXiv:2401.07339},
  eprint = {2401.07339},
  publisher = {arXiv},
  urldate = {2024-02-27},
  archiveprefix = {arXiv},
  langid = {english}
}

@inproceedings{zhang2024gamma,
  title = {Gamma: {{Revisiting Template-based Automated Program Repair}} via {{Mask Prediction}}},
  shorttitle = {Gamma},
  booktitle = {Proceedings of the 38th {{IEEE}}/{{ACM International Conference}} on {{Automated Software Engineering}}},
  author = {Zhang, Quanjun and Fang, Chunrong and Zhang, Tongke and Yu, Bowen and Sun, Weisong and Chen, Zhenyu},
  year = {2024},
  month = sep,
  series = {{{ASE}} '23},
  pages = {535--547},
  publisher = {IEEE Press},
  address = {Echternach, Luxembourg},
  doi = {10.1109/ASE56229.2023.00063},
  urldate = {2025-03-14},
  isbn = {979-8-3503-2996-4}
}

@article{zhang2025patch,
  title = {{{PATCH}}: {{Empowering Large Language Model}} with {{Programmer-Intent Guidance}} and {{Collaborative-Behavior Simulation}} for {{Automatic Bug Fixing}}},
  shorttitle = {{{PATCH}}},
  author = {Zhang, Yuwei and Jin, Zhi and Xing, Ying and Li, Ge and Liu, Fang and Zhu, Jiaxin and Dou, Wensheng and Wei, Jun},
  year = {2025},
  month = feb,
  journal = {ACM Trans. Softw. Eng. Methodol.},
  issn = {1049-331X},
  doi = {10.1145/3718739},
  urldate = {2025-03-15}
}

@inproceedings{zheng2023outline,
  title = {Outline, {{Then Details}}: {{Syntactically Guided Coarse-To-Fine Code Generation}}},
  shorttitle = {Outline, {{Then Details}}},
  booktitle = {Proceedings of the 40th {{International Conference}} on {{Machine Learning}}},
  author = {Zheng, Wenqing and Sharan, S. P. and Jaiswal, Ajay Kumar and Wang, Kevin and Xi, Yihan and Xu, Dejia and Wang, Zhangyang},
  year = {2023},
  month = jul,
  pages = {42403--42419},
  publisher = {PMLR},
  issn = {2640-3498},
  urldate = {2023-09-20},
  langid = {english}
}

@misc{zheng2024opencodeinterpreter,
  title = {{{OpenCodeInterpreter}}: {{Integrating Code Generation}} with {{Execution}} and {{Refinement}}},
  shorttitle = {{{OpenCodeInterpreter}}},
  author = {Zheng, Tianyu and Zhang, Ge and Shen, Tianhao and Liu, Xueling and Lin, Bill Yuchen and Fu, Jie and Chen, Wenhu and Yue, Xiang},
  year = {2024},
  month = feb,
  number = {arXiv:2402.14658},
  eprint = {2402.14658},
  publisher = {arXiv},
  doi = {10.48550/arXiv.2402.14658},
  urldate = {2024-07-25},
  archiveprefix = {arXiv}
}

@inproceedings{zhong2024debug,
  title = {Debug like a {{Human}}: {{A Large Language Model Debugger}} via {{Verifying Runtime Execution Step}} by {{Step}}},
  shorttitle = {Debug like a {{Human}}},
  booktitle = {Findings of the {{Association}} for {{Computational Linguistics ACL}} 2024},
  author = {Zhong, Li and Wang, Zilong and Shang, Jingbo},
  editor = {Ku, Lun-Wei and Martins, Andre and Srikumar, Vivek},
  year = {2024},
  month = aug,
  pages = {851--870},
  publisher = {Association for Computational Linguistics},
  address = {Bangkok, Thailand and virtual meeting},
  urldate = {2024-09-18}
}

@inproceedings{shao2025llms,
  title={Are LLMs Correctly Integrated into Software Systems?},
  author={Shao, Yuchen and Huang, Yuheng and Shen, Jiawei and Ma, Lei and Su, Ting and Wan, Chengcheng},
  booktitle={2025 IEEE/ACM 47th International Conference on Software Engineering (ICSE)},
  pages={1178--1190},
  year={2025},
  organization={IEEE}
}

@inproceedings{zhu2023tare,
  title = {Tare: {{Type-Aware Neural Program Repair}}},
  shorttitle = {Tare},
  booktitle = {2023 {{IEEE}}/{{ACM}} 45th {{International Conference}} on {{Software Engineering}} ({{ICSE}})},
  author = {Zhu, Qihao and Sun, Zeyu and Zhang, Wenjie and Xiong, Yingfei and Zhang, Lu},
  year = {2023},
  month = may,
  pages = {1443--1455},
  issn = {1558-1225},
  doi = {10.1109/ICSE48619.2023.00126},
  urldate = {2024-11-15}
}

@misc{zhu2024deepseekcoderv2,
  title = {{{DeepSeek-Coder-V2}}: {{Breaking}} the {{Barrier}} of {{Closed-Source Models}} in {{Code Intelligence}}},
  shorttitle = {{{DeepSeek-Coder-V2}}},
  author = {DeepSeek AI},
  year = {2024},
  month = jun,
  number = {arXiv:2406.11931},
  eprint = {2406.11931},
  publisher = {arXiv},
  urldate = {2024-06-19},
  archiveprefix = {arXiv},
  langid = {english}
}

@misc{ziegler2020finetuning,
  title = {Fine-{{Tuning Language Models}} from {{Human Preferences}}},
  author = {Ziegler, Daniel M. and Stiennon, Nisan and Wu, Jeffrey and Brown, Tom B. and Radford, Alec and Amodei, Dario and Christiano, Paul and Irving, Geoffrey},
  year = {2020},
  month = jan,
  number = {arXiv:1909.08593},
  eprint = {1909.08593},
  publisher = {arXiv},
  urldate = {2024-01-08},
  archiveprefix = {arXiv},
  langid = {english}
}

@article{wang2025dissect,
  title={Dissect-and-Restore: AI-based Code Verification with Transient Refactoring},
  author={Wang, Changjie and Scazzariello, Mariano and Alshnakat, Anoud and Guanciale, Roberto and Kosti{\'c}, Dejan and Chiesa, Marco},
  journal={arXiv preprint arXiv:2510.25406},
  year={2025}
}

@article{shi2024between,
  title={Between Lines of Code: Unraveling the Distinct Patterns of Machine and Human Programmers. In 2025 IEEE/ACM 47th International Conference on Software Engineering (ICSE)},
  author={Shi, Yuling and Zhang, Hongyu and Wan, Chengcheng and Gu, Xiaodong},
  journal={IEEE Computer Society},
  pages={51--62},
  year={2024}
}

\end{document}